\pdfoutput=1

\documentclass[11pt]{article}

\usepackage[final]{acl}

\usepackage{times}
\usepackage{latexsym}
\usepackage{listings}

\usepackage[T1]{fontenc}

\usepackage[utf8]{inputenc}

\usepackage{microtype}

\usepackage{inconsolata}

\usepackage{graphicx}
\usepackage[table,x11names,dvipsnames]{xcolor}
\usepackage{utfsym}
\usepackage{booktabs}
\usepackage{multirow}

\title{BiMediX2: Bio-Medical EXpert LMM for Diverse Medical Modalities}

\author{Sahal Shaji Mullappilly$^1$\thanks{Equal Contribution}, ~Mohammed Irfan Kurpath$^1$\footnotemark[1], ~Sara Pieri$^1$\\ \textbf{~Saeed Yahya Alseiari$^5$, ~Shanavas Cholakkal$^6$, ~Khaled Aldahmani$^{3,4}$, ~Fahad Khan$^{1,2}$}\\\textbf{~Rao Anwer$^1$, ~Salman Khan$^1$, Timothy Baldwin$^1$, ~Hisham Cholakkal$^1$} \vspace{1mm} \\ 
$^1$Mohamed Bin Zayed University of Artificial Intelligence (MBZUAI), $^2$Linköping University\\ $^3$Shaikh Tahnoon bin Mohammed Medical City (STMC), $^4$Tawam Hospital \\$^5$Sheikh Shakhbout Medical City (SSMC), $^6$Govt Medical College Kozhikode 
}

\begin{document}
\maketitle
\begin{abstract}
We introduce BiMediX2, a bilingual (Arabic-English) Bio-Medical EXpert Large Multimodal Model that supports text-based and  image-based medical interactions. It enables multi-turn conversation in Arabic and English and supports diverse medical imaging modalities, including radiology, CT, and histology. To train BiMediX2, we curate BiMed-V, an extensive Arabic-English bilingual healthcare dataset consisting of 1.6M samples of diverse medical interactions. This dataset supports a range of medical Large Language Model (LLM) and Large Multimodal Model (LMM) tasks, including multi-turn medical conversations, report generation, and visual question answering (VQA). We also introduce BiMed-MBench, the first Arabic-English medical LMM evaluation benchmark, verified by medical experts. BiMediX2 demonstrates excellent performance across multiple medical LLM and LMM benchmarks, achieving state-of-the-art results compared to other open-sourced models. On BiMed-MBench, BiMediX2 outperforms existing methods by over 9\% in English and more than 20\% in Arabic evaluations. Additionally, it surpasses GPT-4 by approximately 9\% in UPHILL factual accuracy evaluations and excels in various medical VQA, report generation, and report summarization tasks. Our trained models, instruction set, and source code are available at -- {\small\url{https://github.com/mbzuai-oryx/BiMediX2}}

\end{abstract}

\section{Introduction}

\begin{figure}[!t]
  \centering
    \setlength{\belowcaptionskip}{-15pt}
    \includegraphics[width=\linewidth]{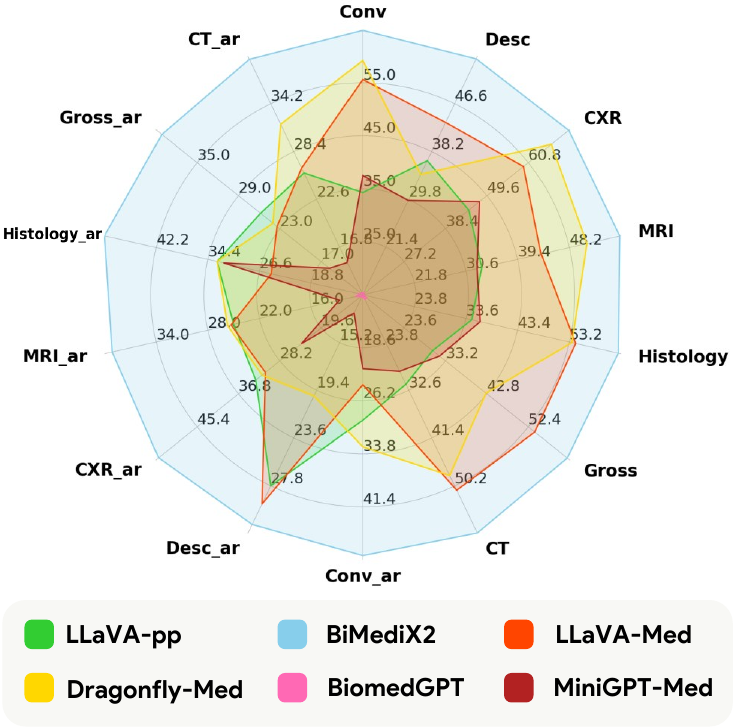}
\caption{\textbf{Performance comparison on BiMed-MBench.} The comparison is conducted across different tasks and modalities, including CT, MRI, CXR, Histology, and Gross, along with their Arabic counterparts (CT\_ar, MRI\_ar, CXR\_ar, Histology\_ar, and Gross\_ar). Each axis represents the performance score for a specific category, highlighting BiMediX2’s superior performance across diverse tasks and modalities in both English and Arabic.}

\label{fig:llavamed_spider}
\end{figure}

\begin{table*}[!t]
    \centering
    \setlength{\belowcaptionskip}{-15pt}
    \resizebox{\textwidth}{!}{
    \begin{tabular}{@{}lccccccccc@{}}
    \toprule
    \multicolumn{1}{c}{\textbf{Model}} & \textbf{MTC} & \textbf{RS} & \textbf{RG} & \textbf{Rad} & \textbf{Oph} & \textbf{Path} & \textbf{Micro} & \textbf{LLM+VLM} & \textbf{Bil (Ar)} \\ \midrule
    \textbf{Meditron \cite{chen2023meditron}} & \textcolor{red}{\usym{2717}} & \textcolor{red}{\usym{2717}} & \textcolor{red}{\usym{2717}} & \textcolor{red}{\usym{2717}} & \textcolor{red}{\usym{2717}} & \textcolor{red}{\usym{2717}} & \textcolor{red}{\usym{2717}} & \textcolor{red}{\usym{2717}} & \textcolor{red}{\usym{2717}} \\
    \textbf{Med42 \cite{christophe2024med42evaluatingfinetuning}} & \textcolor{ForestGreen}{\usym{2713}} & \textcolor{ForestGreen}{\usym{2713}} & \textcolor{red}{\usym{2717}} & \textcolor{red}{\usym{2717}} & \textcolor{red}{\usym{2717}} & \textcolor{red}{\usym{2717}} & \textcolor{red}{\usym{2717}} & \textcolor{red}{\usym{2717}} & \textcolor{red}{\usym{2717}} \\
    \textbf{OpenBioLLM \cite{OpenBioLLMs}} & \textcolor{ForestGreen}{\usym{2713}} & \textcolor{ForestGreen}{\usym{2713}} & \textcolor{red}{\usym{2717}} & \textcolor{red}{\usym{2717}} & \textcolor{red}{\usym{2717}} & \textcolor{red}{\usym{2717}} & \textcolor{red}{\usym{2717}} & \textcolor{red}{\usym{2717}} & \textcolor{red}{\usym{2717}} \\
    \textbf{Llama3.1 \cite{meta-llama3.1}} & \textcolor{ForestGreen}{\usym{2713}} & \textcolor{ForestGreen}{\usym{2713}} & \textcolor{red}{\usym{2717}} & \textcolor{red}{\usym{2717}} & \textcolor{red}{\usym{2717}} & \textcolor{red}{\usym{2717}} & \textcolor{red}{\usym{2717}} & \textcolor{red}{\usym{2717}} & \textcolor{red}{\usym{2717}} \\
    \textbf{BiMediXv1 \cite{pieri2024bimedix}} & \textcolor{ForestGreen}{\usym{2713}} & \textcolor{ForestGreen}{\usym{2713}} & \textcolor{red}{\usym{2717}} & \textcolor{red}{\usym{2717}} & \textcolor{red}{\usym{2717}} & \textcolor{red}{\usym{2717}} & \textcolor{red}{\usym{2717}} & \textcolor{red}{\usym{2717}} & \textcolor{ForestGreen}{\usym{2713}} \\ \bottomrule
    \end{tabular}%
    }
    \vspace{1mm}
    \\
    \resizebox{\textwidth}{!}{
    \begin{tabular}{@{}lcccccccccc@{}}
    \toprule
    \multicolumn{1}{c}{\textbf{Model}} & \textbf{MTC} & \textbf{RS} & \textbf{RG} & \textbf{Rad} & \textbf{Oph} & \textbf{Path} & \textbf{Micro} & \textbf{UM} & \textbf{LLM+VLM} & \textbf{Bil (Ar)} \\ \midrule
    \textbf{LLaVA-pp \cite{hanoona2024LLaVA++}} & \textcolor{ForestGreen}{\usym{2713}} & \textcolor{ForestGreen}{\usym{2713}} & \textcolor{red}{\usym{2717}} & \textcolor{red}{\usym{2717}} & \textcolor{red}{\usym{2717}} & \textcolor{red}{\usym{2717}} & \textcolor{red}{\usym{2717}} & \textcolor{ForestGreen}{\usym{2713}} & \textcolor{red}{\usym{2717}} & \textcolor{red}{\usym{2717}} \\
    \textbf{MiniGPT-Med \cite{alkhaldi2024minigpt}} & \textcolor{red}{\usym{2717}} & \textcolor{ForestGreen}{\usym{2713}} & \textcolor{ForestGreen}{\usym{2713}} & \textcolor{ForestGreen}{\usym{2713}} & \textcolor{red}{\usym{2717}} & \textcolor{red}{\usym{2717}} & \textcolor{red}{\usym{2717}} & \textcolor{ForestGreen}{\usym{2713}} & \textcolor{red}{\usym{2717}} & \textcolor{red}{\usym{2717}} \\
    \textbf{MAIRA-2 \cite{bannur2024maira2groundedradiologyreport}} & \textcolor{red}{\usym{2717}} & \textcolor{red}{\usym{2717}} & \textcolor{ForestGreen}{\usym{2713}} & \textcolor{ForestGreen}{\usym{2713}} & \textcolor{red}{\usym{2717}} & \textcolor{red}{\usym{2717}} & \textcolor{red}{\usym{2717}} & \textcolor{ForestGreen}{\usym{2713}} & \textcolor{red}{\usym{2717}} & \textcolor{red}{\usym{2717}} \\
    \textbf{BioMedGPT \cite{zhang2024generalist}} & \textcolor{red}{\usym{2717}} & \textcolor{ForestGreen}{\usym{2713}} & \textcolor{ForestGreen}{\usym{2713}} & \textcolor{ForestGreen}{\usym{2713}} & \textcolor{ForestGreen}{\usym{2713}} & \textcolor{ForestGreen}{\usym{2713}} & \textcolor{ForestGreen}{\usym{2713}} & \textcolor{red}{\usym{2717}} & \textcolor{red}{\usym{2717}} & \textcolor{red}{\usym{2717}} \\
    \textbf{LLaVA-Med \cite{li2023llavamed}} & \textcolor{ForestGreen}{\usym{2713}} & \textcolor{ForestGreen}{\usym{2713}} & \textcolor{ForestGreen}{\usym{2713}} & \textcolor{ForestGreen}{\usym{2713}} & \textcolor{ForestGreen}{\usym{2713}} & \textcolor{ForestGreen}{\usym{2713}} & \textcolor{ForestGreen}{\usym{2713}} & \textcolor{ForestGreen}{\usym{2713}} & \textcolor{red}{\usym{2717}} & \textcolor{red}{\usym{2717}} \\
    \textbf{Dragonfly-Med \cite{chen2024dragonflymultiresolutionzoomsupercharges}} & \textcolor{red}{\usym{2717}} & \textcolor{ForestGreen}{\usym{2713}} & \textcolor{ForestGreen}{\usym{2713}} & \textcolor{ForestGreen}{\usym{2713}} & \textcolor{ForestGreen}{\usym{2713}} & \textcolor{ForestGreen}{\usym{2713}} & \textcolor{ForestGreen}{\usym{2713}} & \textcolor{ForestGreen}{\usym{2713}} & \textcolor{red}{\usym{2717}} & \textcolor{red}{\usym{2717}} \\
    \rowcolor{blue!10}\textbf{BiMediX2 (ours)} & \textcolor{ForestGreen}{\usym{2713}} & \textcolor{ForestGreen}{\usym{2713}} & \textcolor{ForestGreen}{\usym{2713}} & \textcolor{ForestGreen}{\usym{2713}} & \textcolor{ForestGreen}{\usym{2713}} & \textcolor{ForestGreen}{\usym{2713}} & \textcolor{ForestGreen}{\usym{2713}} & \textcolor{ForestGreen}{\usym{2713}} & \textcolor{ForestGreen}{\usym{2713}} & \textcolor{ForestGreen}{\usym{2713}} \\ \bottomrule
    \end{tabular}%
    }
    \caption{\textbf{Comparison of tasks and modalities addressed by  recent medical LLMs and VLMs. }\textbf{Abbreviations:} \textbf{MTC} (Multi-turn conversation), \textbf{RS} (Report Summarization), \textbf{RG} (Report Generation), \textbf{Rad} (Radiology), \textbf{Oph} (Ophthalmology), \textbf{Path} (Pathology), \textbf{Micro} (Microscopic), \textbf{UM} (Unified Model: Single model checkpoint for all downstream tasks), \textbf{LLM+VLM} (Unified LLM + VLM), \textbf{Bil (Ar)} \mbox{(Bilingual Arabic capabilities).}}

    \label{tab:comparison_llm}
    \end{table*}
    
Recently, medical Large Language Models (LLMs) and medical Large Multimodal Models (LMMs) have shown promising results as conversational assistants for improving accessibility to quality medical advice. However, most medical Vision-Language Models (VLMs) referred to interchangeably as medical LMMs in this paper often compromise their text-based understanding (i.e., medical LLM performance) when integrating multimodal capabilities, making it challenging to interact seamlessly with users (see LLM+VLM column in Tab.~\ref{tab:comparison_llm}). This is particularly challenging when users initially ask general medical queries in text format and later follow up with questions related to user-provided medical images introduced mid-conversation to provide \mbox{additional supporting information.}

In addition to supporting diverse LLM and LMM tasks, such as multi-turn conversations and report generation, it is also desirable to support various medical image modalities, such as radiology and pathology, in a unified model. 
However, as shown in Tab.~\ref{tab:comparison_llm}, state-of-the-art medical LMMs, such as MiniGPT-Med \cite{alkhaldi2024minigpt}, are either restricted to a limited set of medical image modalities, such as radiology, or require separately fine-tuned models for each downstream task (e.g., BiomedGPT \cite{zhang2024generalist}), posing significant challenges for real-world deployment.

Moreover, advancements in medical LLMs and LMMs remain predominantly English-centric, leaving significant gaps for non-English-speaking populations, particularly in languages like Arabic, spoken by over 400 million people. Unlike most state-of-the-art medical LLMs, BiMediX \cite{pieri2024bimedix} (referred to as BiMediXv1 in this paper for clarity) supports Arabic-English bilingual interactions and offers diverse text-based interaction capabilities, such as multi-turn conversations, report summarization, and question-answering. However, it is limited to text modality and lacks medical image understanding capabilities.

To address the aforementioned limitations of existing medical LMMs, we propose BiMediX2, a Bio-Medical EXpert Large Multimodal Model that supports diverse medical tasks and modalities while also facilitating seamless user interactions \mbox{in both English and Arabic.} 
\subsection{Contributions}
\begin{figure*}[t!]
  \centering
    \setlength{\belowcaptionskip}{-10pt}
    \includegraphics[width=\linewidth]{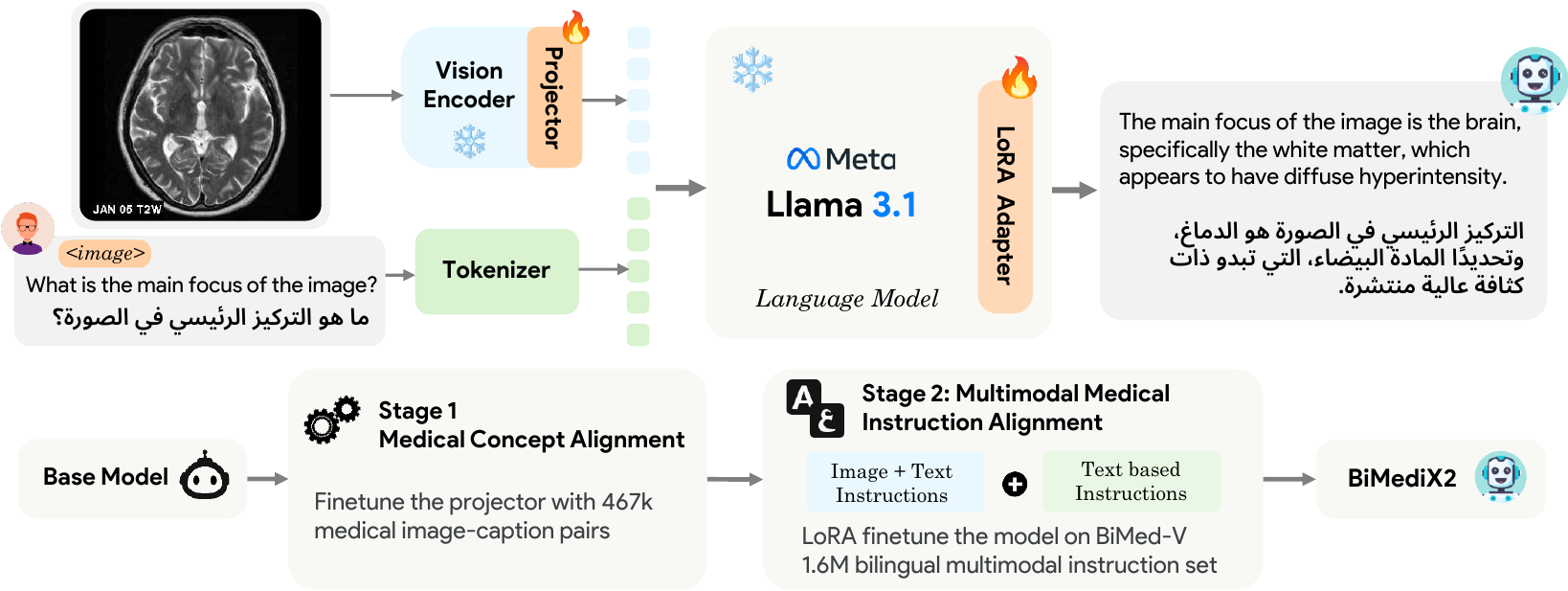}
\caption{\textbf{BiMediX2: Overall Architecture} Our model is designed for medical image analysis and bilingual multi-turn conversations. Medical images are processed through a Vision Encoder and aligned with a Projector, while the text inputs are tokenized using the default tokenizer. The resulting tokens are then passed into the language model (Meta Llama 3.1) to generate responses in the prompted language. We only train the language model using LoRA adapters, while the projector is finetuned for medical image-text alignment. BiMediX2 follows a two-stage training pipeline. \textit{Stage-1} aligns medical visual concepts using 467K image-caption pairs. \textit{Stage-2} performs multimodal medical instruction tuning with our proposed BiMed-V 1.6M bilingual instructions comprising both image-text and text-only medical instructions.}
    \label{fig:main_arch}
\end{figure*}
We introduce  BiMediX2 an Arabic-English bilingual medical LMM that supports  broader spectrum of text-based and multimodal medical tasks, including Multi-Turn Conversations (MTC), Report Summarization (RS), Report Generation (RG), Radiology (Rad), Ophthalmology (Oph), Pathology (Path), Microscopic Analysis (Micro), in a Unified Model (UM) with LLM and VLM capabilities (LLM+VLM) as shown in Tab.~\ref{tab:comparison_llm}.  It supports a wide range of imaging modalities, including Chest X-rays, CT scans, MRIs, Histology slides, and Gross pathology images (see Fig.~\ref{fig:llavamed_spider}).  The key contributions of our work can be  summarized as: 
\noindent\textbf{(i)} We curated \textbf{\textit{a comprehensive Arabic-English  bilingual healthcare specialized instruction set named BiMed-V}} comprising over \textbf{\textit{1.6M instructions }} including text-only and  image-text  instructions across diverse medical image modalities.

\noindent\textbf{(ii)} We introduce \textbf{\textit{the first Arabic-English bilingual medical LMM evaluation benchmark  named BiMed-MBench}}, consisting of 386 medical queries in English and Arabic across various medical image modalities, fully \textbf{\textit{verified by }} medical experts. 

\noindent\textbf{(iii)} We instruction-tune Llama3.1 LLM with our BiMed-V to obtain \textit{ \textbf{the first Arabic-English bilingual medical LMM}}  achieving state-of-the-art results among open-sourced models on various VLM evaluation benchmarks across diverse medical image modalities while also excelling on medical LLM evaluation benchmarks \textbf{\textit{in both English and Arabic}}. This is achieved through our multistage training pipeline and carefully curated instruction set, which balances text-only and multimodal \mbox{medical instructions.}

\noindent\textbf{(iv)} We perform \textbf{\textit{comprehensive evaluation of our model across 12 diverse evaluation benchmarks}}. In addition to achieving promising results on our bilingual BiMed-MBench, BiMediX2 achieves \textbf{\textit{excellent results for visual question answering}} on Path-VQA, SLAKE and Rad-VQA  evaluation benchmarks, and \textbf{\textit{report generation task}} on MIMIC CXR  and \textbf{\textit{report summarization }}on MIMIC-III benchmarks. Additionally, BiMediX2 achieves \textbf{\textit{favorable results on diverse medical LLM benchmarks}}, such  Medical MMLU, MedMCQA, MedQA, USMLE, and PubMedQA datasets, and it also shows \textbf{\textit{robustness in discerning and correcting misinformation in medical }}context on the UPHILL OpenQA Evaluation.

Fig~\ref{fig:llavamed_spider} illustrates the performance of SOTA medical LMMs on our BiMed-MBench evaluation. BiMediX2 achieves SOTA results on BiMed-MBench, with over a 9\% improvement in English evaluations and more than a 20\% improvement in Arabic evaluations. Furthermore, it excels in medical Visual Question Answering, Report Generation, and Report Summarization tasks. Similarly, BiMediX2 outperforms GPT-4 by more than 8\% on the USMLE benchmark and by more than 9\% in UPHILL factual accuracy evaluations.

\section{BiMediX2}

\begin{table*}[t!]
\setlength{\belowcaptionskip}{-10pt}
\resizebox{\textwidth}{!}{%
\begin{tabular}{@{}lccccccccccc@{}}
\toprule
\multicolumn{1}{c}{\multirow{2}{*}{\textbf{Model}}} & \multicolumn{6}{c}{\textbf{MMLU}} & \multirow{2}{*}{\textbf{MedMCQA}} & \multirow{2}{*}{\textbf{MedQA}} & \multirow{2}{*}{\textbf{USMLE}} & \multirow{2}{*}{\textbf{PubmedQA}} & \multirow{2}{*}{\textbf{Average}} \\ \cmidrule(lr){2-7}
\multicolumn{1}{c}{} & \textbf{Cli-KG} & \textbf{C-Bio} & \textbf{C-Med} & \textbf{Med-Gen} & \textbf{Pro-Med} & \textbf{Ana} &  &  &  &  &  \\ \midrule
\textbf{BioMedGPT-LM-7B} & 49.4 & 43.1 & 41.4 & 45.0 & 51.0 & 45.2 & 34.8 & 33.2 & 31.7 & 74.0 & 44.9 \\
\textbf{LLaVA-Med} & 59.6 & 59.7 & 50.9 & 59.0 & 51.5 & 51.9 & 44.5 & 35.7 & 36.9 & 74.0 & 52.4 \\
\textbf{Dragonfly-Med} & 65.6 & 69.4 & 56.6 & 69.0 & 58.4 & 57.0 & 49.9 & 42.8 & 46.1 & 75.4 & 59.0 \\
\textbf{Apollo-7B} & 64.2 & 73.6 & 59.5 & 70.0 & 70.6 & 61.5 & 54.4 & 50.2 & 52.3 & 39.0 & 59.5 \\
\textbf{GPT 3.5} & 69.8 & 72.2 & 61.3 & 70.0 & 70.2 & 56.3 & 50.1 & 50.8 & 49.1 & 71.6 & 62.1 \\
\textbf{Meditron 70B} & 68.3 & 77.8 & 63.6 & 75.0 & 74.6 & 56.3 & 48.4 & 53.1 & 55.4 & 76.2 & 64.9 \\
\textbf{BiMediXv1} & 78.9 & 86.1 & 68.2 & 85.0 & 80.5 & 74.1 & 62.7 & 62.8 & 66.8 & 80.2 & 74.5 \\
\textbf{Apollo-72B} & 82.3 & 90.3 & 77.5 & 85.0 & 86.0 & 70.4 & 66.7 & 65.3 & 74.2 & 78.8 & 77.6 \\
\textbf{GPT 4} & 86.0 & 95.1 & 76.9 & 91.0 & 93.0 & 80.0 & 69.5 & 78.9 & 83.8 & 75.2 & 82.9 \\
\textbf{Llama3-Med42-70B} & 84.2 & 93.1 & 79.8 & 91.0 & 90.1 & 80.7 & 72.5 & 73.8 & 84.3 & 80.6 & 83.0 \\
\textbf{OpenBioLLM-70B} & 92.5 & 93.8 & 85.6 & 93.0 & 93.4 & 83.7 & 74.1 & 68.9 & 72.0 & 78.0 & 83.5 \\
\textbf{Llama 3.1 70B} & 83.4 & 95.1 & 79.2 & 93.0 & 91.5 & 80.7 & 71.7 & 73.8 & 92.0 & 77.6 & 83.8 \\ \midrule
\rowcolor{blue!10} \textbf{BiMediX2 4B} & 55.1 & 63.9 & 47.4 & 55.0 & 36.0 & 52.6 & 38.1 & 37.9 & 47.1 & 72.2 & 50.5 \\
\rowcolor{blue!10} \textbf{BiMediX2 8B} & 77.7 & 79.2 & 68.8 & 82.0 & 74.3 & 65.9 & 58.0 & 57.0 & 68.6 & 72.4 & 70.4 \\
\rowcolor{blue!10} \textbf{BiMediX2 70B} & \textbf{86.8} & \textbf{95.1} & \textbf{79.8} & \textbf{94.0} & \textbf{91.5} & \textbf{82.2} & \textbf{70.5} & \textbf{74.3} & \textbf{92.3} & \textbf{79.0} & \textbf{84.6} \\ \bottomrule
\end{tabular}%
}
\caption{\textbf{Clinical LLM Evaluation Benchmark}}
\label{tab:lmeval_sota}
\end{table*}

The architecture of BiMediX2 is designed to facilitate seamless integration of medical image analysis and bilingual multi-turn conversations. At its core, (see Fig~\ref{fig:main_arch}) the model employs a Vision Encoder \cite{radford2021learningtransferablevisualmodels} to process a diverse array of medical imaging modalities, including chest X-rays, CT scans, MRIs, histology slides, and gross pathology images. This visual data is aligned with textual inputs through a dedicated Projector, ensuring accurate and contextually rich medical image-text mapping following \cite{liu2023visualinstructiontuning}. As shown in Fig~\ref{fig:main_arch} we use the `$<$\textit{image}$>$' token as a place holder to encode the visual features for multimodal medical instructions. For text based medical data the inputs are processed using a standard tokenizer, transforming them into the language embedding space of Llama 3.1 \cite{meta-llama3.1}. This design enables BiMediX2 to generate precise and context-aware responses in either English or Arabic, depending on the user prompt and supports multimodal interactions while preserving the medical LLM capabilities.

Key to BiMediX2’s performance is its modular and efficient training approach. LoRA adapters \cite{hu2021lora} are utilized to fine-tune the language model while maintaining computational efficiency and minimizing resource demands. The projector is simultaneously fine-tuned to optimize image-text alignment in a medical context. Furthermore, the system is supported by a robust data generation framework, where a comprehensive English data corpus is translated into Arabic using GPT-4o. A random subset of this translation is meticulously verified by bilingual medical experts to ensure clinical relevance and linguistic accuracy. This pipeline enables BiMediX2 to excel in medical tasks, including report generation, radiology analysis, pathology insights, and ophthalmological assessments, in a unified, bilingual, and multimodal framework.

\subsection{BiMed-V: Multimodal, Bilingual Dataset}

The BiMed-V dataset is a comprehensive bilingual and multimodal instruction set comprising of 1.6M samples, developed to enhance medical image-text alignment and multimodal understanding. It incorporates a diverse range of publicly available datasets, such as PMC-OA \cite{lin2023pmc}, Rad-VQA \cite{lau2018dataset}, Path-VQA \cite{he2020pathological} and SLAKE \cite{liu2021slake} complemented by custom-curated data. We also curated 163k VQA samples by repurposing the LLaVA-Med \cite{li2023llavamed} 60K-IM dataset. Furthermore, over 10k samples from the LLaVA-Med pretraining dataset were reformatted into interactive conversations using the Llama 3.1 70B model. A subset of the PMC-OA dataset with short question-answer pairs and multiple-choice questions were added to enhance the dataset's diversity. Training splits of Rad-VQA, Path-VQA and SLAKE, which typically feature concise answers, were restructured into more detailed responses using the same Llama 3.1 70B model, enhancing the dataset’s depth and usability for complex tasks.

A unique feature of BiMed-V is its bilingual support, facilitated by a multimodal open-ended instruction set comprising 326k samples across various medical imaging modalities. This includes 163k Arabic-language samples generated via a comprehensive translation framework (see Fig.~\ref{fig:data_translation}). English datasets were translated into Arabic using GPT-4o, with  verification of a random subset by bilingual medical experts to ensure clinical relevance and linguistic precision. This hybrid approach balances automation and expert validation, significantly reducing reliance on human medical domain experts while maintaining data quality (Please refer \ref{sec:data_translation} for more details on translation framework and expert validation). Additionally, the inclusion of text-based clinical data from BiMediXv1 \cite{pieri2024bimedix} ensures the dataset retains robust language understanding capabilities while expanding its multimodal medical proficiency. This extensive dataset forms the foundation for advanced medical image-text alignment and conversational multimodal applications. For dataset composition, see Section \ref{sec:dataset_comp}.

\begin{table*}[!t]
\centering
\resizebox{0.8\textwidth}{!}{%
\begin{tabular}{@{}lcccccccc@{}}
\toprule
\multicolumn{1}{c}{\textbf{Model}} & \textbf{Conversation} & \textbf{Description} & \textbf{CXR} & \textbf{MRI} & \textbf{Histology} & \textbf{Gross} & \textbf{CT} & \textbf{Overall} \\ \midrule
\textbf{BiomedGPT} & 15.3 & 13.3 & 16.4 & 13.0 & 14.1 & 14.9 & 15.8 & 14.8 \\
\textbf{MAIRA-2} & 19.1 & 27.0 & 45.6 & 13.7 & 13.8 & 15.0 & 19.1 & 21.2 \\
\textbf{LLaVA-pp} & 34.3 & 36.6 & 44.7 & 33.3 & 34.7 & 30.2 & 31.5 & 34.9 \\
\textbf{MiniGPT-Med} & 37.5 & 29.6 & 47.6 & 32.5 & 36.3 & 31.8 & 29.1 & 35.4 \\
\textbf{LLaVA-Med} & 55.6 & 43.3 & 59.5 & 43.4 & 54.4 & 53.9 & 51.0 & 52.4 \\
\textbf{Dragonfly-Med} & 59.2 & 34.2 & 67.0 & 51.2 & 53.7 & 42.6 & 48.3 & 52.7 \\
\rowcolor{blue!10}\textbf{BiMediX2 8B} & \textbf{64.9} & \textbf{54.5} & \textbf{71.7} & \textbf{56.8} & \textbf{62.5} & \textbf{61.4} & \textbf{58.9} & \textbf{62.2} \\ \bottomrule
\end{tabular}%
    }
\caption{\textbf{BiMed-MBench English Evaluation}}
\label{tab:llavamed_eval_eng}
\end{table*}

\begin{table*}[!t]
\centering
\resizebox{0.8\textwidth}{!}{%
\begin{tabular}{@{}lcccccccc@{}}
\toprule
\multicolumn{1}{c}{\textbf{Model}} & \textbf{Conversation} & \textbf{Description} & \textbf{CXR} & \textbf{MRI} & \textbf{Histology} & \textbf{Gross} & \textbf{CT} & \textbf{Overall} \\ \midrule
\textbf{BiomedGPT} & 11.1 & 11.2 & 11.4 & 10.8 & 11.5 & 11.3 & 11.1 & 11.2 \\
\textbf{MAIRA-2} & 14.0 & 12.5 & 25.6 & 10.6 & 12.8 & 11.6 & 12.7 & 14.6 \\
\textbf{MiniGPT-Med} & 21.6 & 12.6 & 23.7 & 12.7 & 32.0 & 15.8 & 14.9 & 20.2 \\
\textbf{LLaVA-Med} & 23.9 & 29.4 & 31.2 & 25.3 & 24.8 & 23.4 & 26.4 & 26.2 \\
\textbf{LLaVA-pp} & 29.0 & 27.8 & 33.2 & 25.0 & 33.0 & 25.8 & 25.8 & 28.7 \\
\textbf{Dragonfly-Med} & 32.8 & 19.9 & 31.9 & 25.7 & 33.0 & 24.0 & 31.7 & 29.5 \\
\rowcolor{blue!10}\textbf{BiMediX2 8B} & \textbf{54.3} & \textbf{36.2} & \textbf{61.4} & \textbf{44.6} & \textbf{51.5} & \textbf{43.5} & \textbf{50.8} & \textbf{50.5} \\ \bottomrule
\end{tabular}%
    }
\caption{\textbf{BiMed-MBench Arabic Evaluation}}
\label{tab:llavamed_eval_ara}
\end{table*}

\subsection{Medical Instruction Tuning}

To enable BiMediX2’s robust capabilities in both bilingual and multimodal medical tasks, we employ a two-stage training process that ensures precise alignment of visual and textual representations while adapting the language model for complex medical instruction tasks (see Fig~\ref{fig:main_arch}).  

\noindent\textbf{Stage 1: Medical Concept Alignment:}
In the first stage, we finetune the Projector alone to align visual embeddings to the language embedding space. The training utilizes a comprehensive dataset of 467k image-caption pairs sourced from the LLaVA-Med \cite{li2023llavamed} pretraining dataset. These pairs span diverse medical imaging modalities and captions that describe clinically relevant features.

\noindent\textbf{Stage 2: Multimodal Medical Instruction Alignment:} 
The second stage finetunes the LoRA\cite{hu2021lora} adapters within the language model, enhancing its ability to process and generate multimodal medical instructions. For this, we utilize our BiMed-V 1.6M bilingual multimodal instruction set, which comprises carefully crafted English and Arabic prompts paired with corresponding visual and textual responses. This dataset enables the model to learn nuanced instructions across a wide array of medical domains, from radiology to pathology, in a bilingual context. 

Through these two stages, BiMediX2 achieves seamless integration of bilingual and multimodal capabilities, enabling it to deliver accurate and context-aware medical insights in both English and Arabic, tailored to a variety of clinical scenarios. We have used LLaMA 3.1 (8B, 70B) and Phi-3.5 V \cite{abdin2024phi} as base models to obtain BiMediX2 8B, BiMediX2 70B, and BiMediX2 4B variants, respectively. For further details on model and training configurations, see Section \ref{sec:model_config}.

\section{Experiments}

In the literature, evaluating medical language models predominantly involves multiple-choice question-answering tasks, with accuracy as the performance metric. We employed the EleutherAI \cite{eval-harness} evaluation framework for evaluating text-only medical tasks following \cite{pieri2024bimedix}. For our bilingual and multimodal evaluations, we utilized several benchmarks and tools. We leveraged our BiMed-MBench for assessing the correctness, helpfulness, relevance and accuracy of the model in a bilingual medical multimodal context. Additionally, we evaluated the model on VQA using datasets such as Path-VQA, SLAKE, and Rad-VQA. For report generation, we used the MIMIC-CXR \cite{johnson2019mimic} dataset, and for report summarization, we used the MIMIC-III \cite{johnson2016mimic} dataset. These evaluations were conducted using the MultiMedEval \cite{royer2024multimedeval} toolkit, ensuring a comprehensive assessment of the model's multimodal capabilities.

\subsection{Medical LLM Benchmarks}

We assess our models on text-based clinical benchmarks using PubMedQA, MedMCQA, MedQA, USMLE, Medical MMLU, and UPHILL OpenQA. For further details, see Section~\ref{sec:llm_bench}.

\begin{figure*}[!t]
  \centering
    \includegraphics[width=\linewidth]{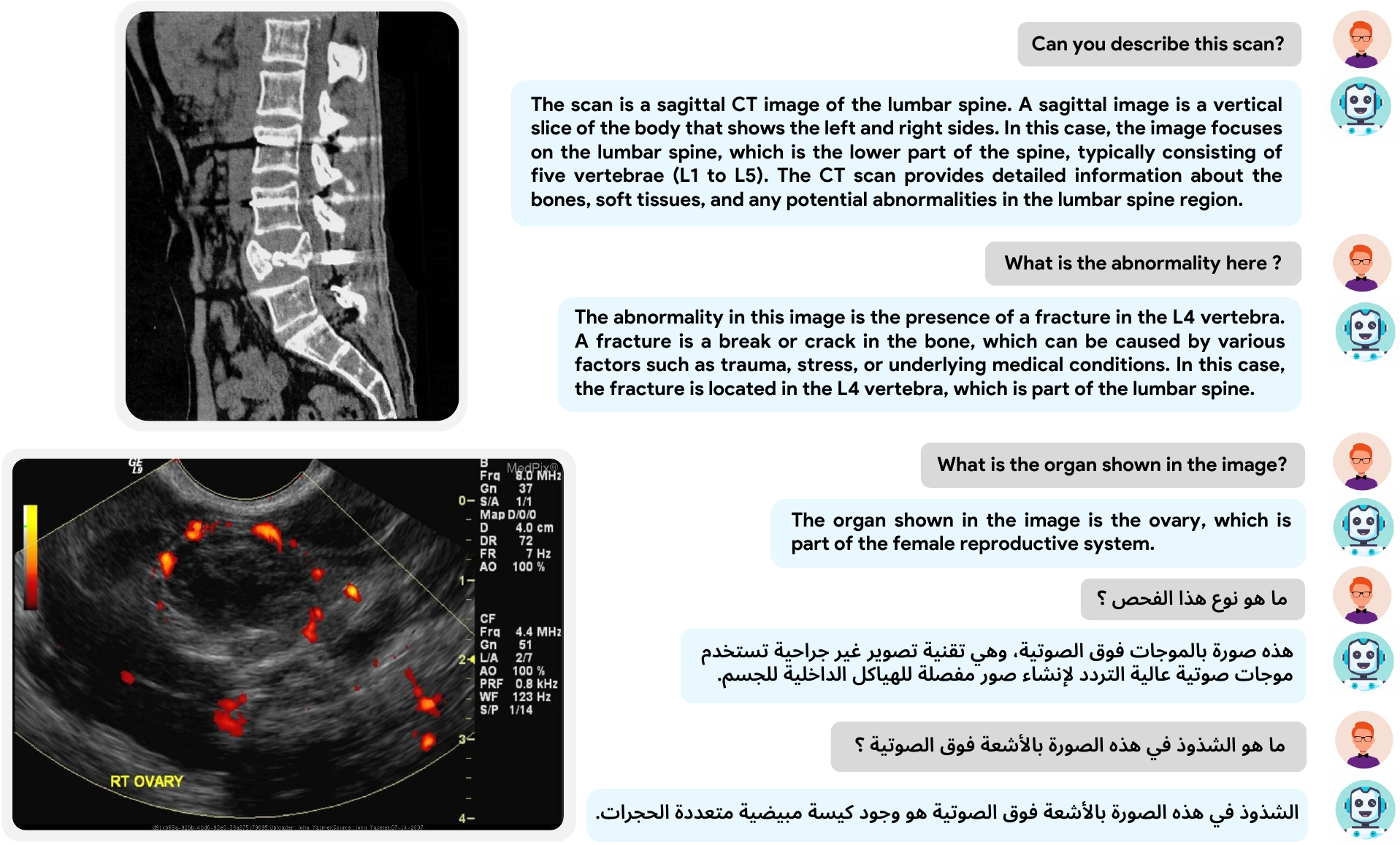}
    \caption{\textbf{Qualitative Examples of BiMediX2 for Medical Image Understanding in a Conversational Context.}}
    \label{fig:main_qualitative}
\end{figure*}

\begin{figure}[!t]
  \centering
    \setlength{\belowcaptionskip}{-15pt}
    \includegraphics[width=0.99\linewidth]{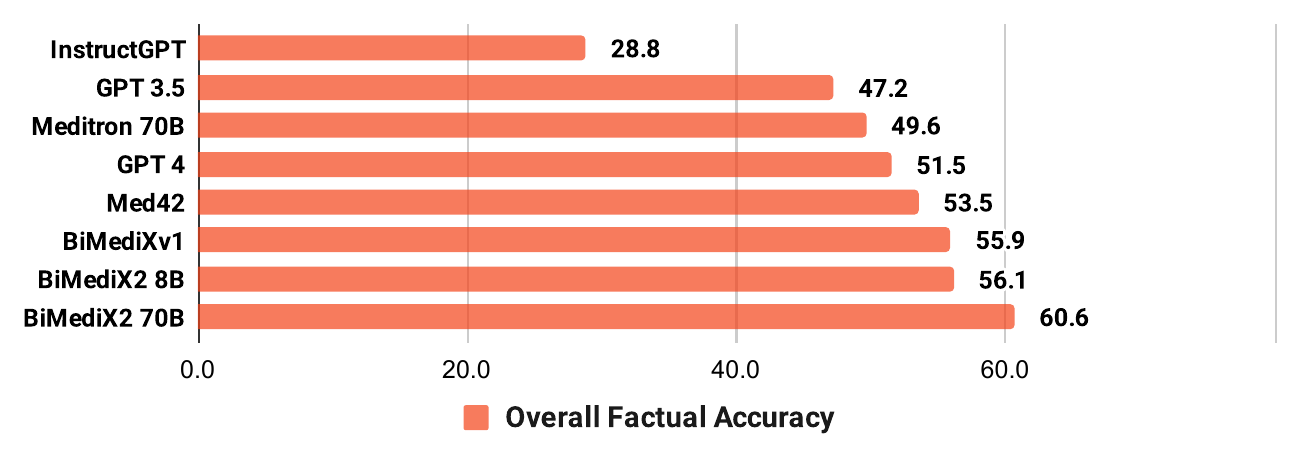}
    \caption{Performance comparison on \textbf{UPHILL OpenQA} \cite{kaur2023evaluating}, assessing the model's ability to address false medical claims at different \mbox{presupposition levels.}}
    \label{fig: uphill_eval}
\end{figure}

\subsection{Medical VLM Benchmarks}

\textbf{BiMed-MBench:} We introduced the first bilingual GPT-4o-based medical LMM benchmark, consisting of 386 medical queries spanning various medical imaging modalities. The test dataset includes conversational interactions and detailed descriptions for modalities such as chest X-rays (CXR), MRI, histology, gross and CT scans derived from LLaVA-Med \cite{li2023llavamed}. GPT-4o evaluates the correctness of model responses based on the provided image context and caption. The reference prediction, serving as the upper-bound answer for the teacher model, is generated using GPT-4 \cite{achiam2023gpt}. For Arabic evaluations, the ground truth is derived from the upper-bound, translated by GPT-4o, and all test samples are comprehensively verified by medical experts for quality assurance. GPT-4o evaluates the responses by comparing the candidate LMM against the upper-bound for English queries and against the Arabic ground truth for Arabic queries. The evaluation assesses the helpfulness, relevance, accuracy, and level of detail of the responses while maintaining the user language. The scoring is conducted on a scale of 1 to 10, with higher scores indicating better performance. Additionally, GPT-4o provides detailed explanations of the evaluation to offer deeper insights into the models' performance. Finally, we compute and compare the relative scores of \mbox{the two candidates.}

\noindent\textbf{Visual Question Answering (VQA):} To evaluate a model's performance in Visual Question Answering, three datasets are used: Path-VQA, SLAKE, and Rad-VQA. These datasets include a mix of open-ended and close-ended questions, designed to test the model's ability to interpret and respond to queries based on medical images. The evaluation employs accuracy, precision, recall, and F1 score to measure the correctness and completeness of the model’s answers, which are critical for clinical reliability. Additionally, BLEU ensures the generated answers are naturally phrased and easy to understand for clinical communication.

\begin{table*}[!t]
\centering
\resizebox{\textwidth}{!}{%
\begin{tabular}{@{}clccccccc@{}}
\toprule
\textbf{Dataset} & \multicolumn{1}{c}{\textbf{Metric}} & \textbf{RadFM} & \textbf{LLaVA Med} & \textbf{BioMedGPT} & \textbf{MiniGPT-Med} & \textbf{Phi-3.5 V} & \textbf{BiMediX2 4B} & \textbf{BiMediX2 8B} \\ \midrule
\multirow{6}{*}{\textbf{Rad-VQA}} & BLEU-1↑ & 0.475 & 0.033 & 0.044 & 0.662 & 0.377 & 0.501 & 0.552 \\
 & closed Q accuracy↑ & 0.577 & 0.545 & 0.203 & 0.829 & 0.618 & 0.685 & 0.725 \\
 & open Q recall↑ & 0.407 & 0.246 & 0.199 & 0.546 & 0.295 & 0.292 & 0.363 \\
 & recall↑ & 0.438 & 0.372 & 0.199 & 0.703 & 0.475 & 0.511 & 0.565 \\
 & open Q accuracy↑ & 0.335 & 0.140 & 0.150 & 0.490 & 0.200 & 0.225 & 0.305 \\
 & F1 ↑ & 0.442 & 0.069 & 0.064 & 0.675 & 0.391 & 0.516 & 0.569 \\ \midrule
\multirow{6}{*}{\textbf{Slake-VQA}} & BLEU-1↑ & 0.746 & 0.036 & 0.175 & 0.337 & 0.089 & 0.625 & 0.778 \\
 & closed Q accuracy↑ & 0.752 & 0.512 & 0.248 & 0.572 & 0.535 & 0.744 & 0.831 \\
 & open Q recall↑ & 0.758 & 0.429 & 0.293 & 0.308 & 0.377 & 0.624 & 0.763 \\
 & recall↑ & 0.695 & 0.443 & 0.260 & 0.396 & 0.404 & 0.664 & 0.786 \\
 & open Q accuracy↑ & 0.725 & 0.362 & 0.259 & 0.278 & 0.329 & 0.567 & 0.729 \\
 & F1 ↑ & 0.714 & 0.075 & 0.192 & 0.349 & 0.129 & 0.641 & 0.787 \\ \midrule
\multirow{6}{*}{\textbf{Path-VQA}} & BLEU-1↑ & 0.257 & 0.021 & 0.145 & 0.296 & 0.283 & 0.469 & 0.587 \\
 & closed Q accuracy↑ & 0.505 & 0.512 & 0.260 & 0.581 & 0.553 & 0.708 & 0.872 \\
 & open Q recall↑ & 0.020 & 0.116 & 0.093 & 0.040 & 0.063 & 0.239 & 0.314 \\
 & recall↑ & 0.221 & 0.287 & 0.176 & 0.311 & 0.308 & 0.474 & 0.593 \\
 & open Q accuracy↑ & 0.005 & 0.053 & 0.077 & 0.019 & 0.027 & 0.210 & 0.282 \\
 & F1 ↑ & 0.232 & 0.052 & 0.154 & 0.299 & 0.287 & 0.475 & 0.595 \\ \midrule
\rowcolor{yellow!20}\multicolumn{2}{c}{\textbf{Average}} & \textbf{0.461} & \textbf{0.239} & \textbf{0.177} & \textbf{0.427} & \textbf{0.319} & \textbf{0.509} & \textbf{0.611} \\ \bottomrule
\end{tabular}%
}
\caption{\textbf{Medical VQA Benchmark} MultiMedEval~\cite{royer2024multimedeval}}
\label{tab:vqa}
\end{table*}

\noindent\textbf{Report Generation:} The MIMIC-CXR \cite{johnson2019mimic} dataset, which includes de-identified radiology reports associated with chest X-rays, is utilized to assess the model's performance in generating the findings section of medical reports. For this task, the input consists of one or more radiology images related to a case, followed by a prompt such as, ``$<$\textit{image}$>$ $<$\textit{image}$>$ \textit{Please caption this scan with findings and impressions.}" This setup evaluates the model's ability to generate accurate and coherent medical reports based \mbox{on visual inputs.}

\noindent\textbf{Report Summarization:} We conduct an evaluation of report summarization using the MIMIC-III \cite{johnson2016mimic} dataset. The task involves generating the impressions section of a radiology report based on the findings section. To achieve this, the model is provided with the free-text findings along with a task prompt, \textit{``Summarize the findings"}. This assessment focuses on the model's ability to distill comprehensive medical information into a concise and precise summary, which is essential for clear and effective \mbox{medical communication.}

In both report generation and summarization tasks, relying solely on lexical metrics such as BLEU, ROUGE-L, and METEOR is insufficient, as these do not guarantee clinical accuracy. Therefore, we complement them with clinically-informed metrics such as F1-RadGraph, CheXbert similarity, and RadCliQ. These metrics assess the preservation of key medical entities, relations, and overall clinical correctness, ensuring that the generated or summarized content aligns not just in phrasing but also in diagnostic intent and clinical relevance.

\begin{table*}[!t]
\centering
\resizebox{0.8\textwidth}{!}{%
\begin{tabular}{@{}clccccc@{}}
\toprule
\textbf{Dataset} & \multicolumn{1}{c}{\textbf{Metric}} & \textbf{LLaVA Med} & \textbf{Dragonfly-Med} & \textbf{BiMediX2 4B} & \textbf{BiMediX2 8B} & \textbf{Med-PaLM M} \\ \midrule
\multirow{7}{*}{\textbf{MIMIC-III}} & ROUGE-L↑ & 0.185 & 0.072 & 0.209 & 0.205 & 0.320 \\
 & BLEU-1↑ & 0.192 & 0.062 & 0.153 & 0.178 & 0.154 \\
 & BLEU-4↑* & 0.520 & 0.000 & 0.410 & 0.449 & - \\
 & F1-RadGraph↑ & 0.232 & 0.000 & 0.222 & 0.230 & 0.347 \\
 & RadCliQ↑* & 0.753 & 0.247 & 0.923 & 0.918 & - \\
 & CheXbert vector↑ & 0.600 & 0.326 & 0.633 & 0.593 & - \\
 & METEOR↑ & 0.303 & 0.060 & 0.264 & 0.339 & - \\ \midrule
\rowcolor{yellow!20}\multicolumn{2}{c}{\textbf{Average}} & \textbf{0.398} & \textbf{0.110} & \textbf{0.402} & \textbf{0.416} & - \\ \bottomrule
\end{tabular}%
}
\caption{\textbf{Report Summarization} MultiMedEval~\cite{royer2024multimedeval}}
\label{tab:report_summary}
\end{table*}

\begin{table*}[!t]
\centering
\resizebox{\textwidth}{!}{%
\begin{tabular}{@{}clccccccc@{}}
\toprule
\textbf{Dataset} & \multicolumn{1}{c}{\textbf{Metric}} & \textbf{RadFM} & \textbf{LLaVA Med} & \textbf{BioMedGPT} & \textbf{BiMediX2 4B} & \textbf{BiMediX2 8B} & \textbf{MAIRA-2 †} & \textbf{Med-PaLM M} \\ \midrule
\multirow{7}{*}{\textbf{\begin{tabular}[c]{@{}c@{}}MIMIC-CXR \\ Report \\ Generation\end{tabular}}} & F1-RadGraph↑ & 0.042 & 0.048 & 0.000 & 0.083 & 0.098 & 0.162 & 0.267 \\
 & BLEU-1↑ & 0.006 & 0.163 & 0.003 & 0.046 & 0.155 & 0.148 & 0.323 \\
 & BLEU-4↑* & 0.000 & 0.060 & 0.000 & 0.042 & 0.016 & 0.104 & 0.115 \\
 & ROUGE-L↑ & 0.065 & 0.125 & 0.012 & 0.131 & 0.153 & 0.164 & 0.275 \\
 & RadCliQ↑* & 0.655 & 0.660 & 0.827 & 0.865 & 0.860 & 0.885 & - \\
 & CheXbert vector↑ & 0.197 & 0.150 & 0.153 & 0.205 & 0.189 & 0.333 & - \\
 & METEOR↑ & 0.053 & 0.137 & 0.016 & 0.107 & 0.174 & 0.187 & -  \\ \midrule
\rowcolor{yellow!20}\multicolumn{2}{c}{\textbf{Average}} & \textbf{0.145} & \textbf{0.192} & \textbf{0.145} & \textbf{0.211} & \textbf{0.235} & \textbf{0.283} & - \\ \bottomrule
\end{tabular}%
}
\caption{\textbf{Report Generation} MultiMedEval~\cite{royer2024multimedeval}}
\label{tab:report_generation}
\end{table*}

\section{Results}

\noindent\textbf{LLM Medical Evaluation:} The performance of BiMediX2 and other models on various language-based medical benchmarks is presented in Fig~\ref{fig:lmeval_sota} and Tab~\ref{tab:lmeval_sota}. Our BiMediX2 70B achieved the highest average score of 84.6\%, outperforming other models such as GPT-4 (82.9\%) and Llama-3-Med42-70B (83.0\%). BiMediX2 70B exhibited superior average performance across multiple datasets, including Medical MMLU, MedMCQA, MedQA, USMLE, and PubMedQA, demonstrating its strong understanding of medical contexts. 

\noindent\textbf{UPHILL OpenQA Evaluation:} Fig~\ref{fig: uphill_eval} illustrates the performance comparison on the UPHILL OpenQA benchmark. BiMediX2 70B achieved the highest overall factual accuracy of 60.6\%, and the second highest being BiMediX2 8B (56.1\%), surpassing other models such as GPT-4 (51.5\%), Meditron 70B (49.6\%), and Med42 (53.5\%). This highlights BiMediX2's effectiveness in discerning and correcting misinformation in medical contexts.

\noindent\textbf{BiMed-MBench Evaluation:} Tables~\ref{tab:llavamed_eval_eng} and~\ref{tab:llavamed_eval_ara} present the evaluation results of BiMediX2 and other medical LMMs on the English and Arabic BiMed-MBench benchmark, respectively. BiMediX2 8B demonstrated superior performance in both English (overall score of 62.2\%) and Arabic (overall score of 50.5\%) evaluations, outperforming other models. Fig~\ref{fig:llavamed_spider} compares the performance of state-of-the-art medical LMMs on our BiMed-MBench evaluation in a bilingual context.  
This indicates BiMediX2's strong bilingual capabilities and its effectiveness in handling medical conversations and descriptions across \mbox{different imaging modalities.}

\noindent\textbf{Medical VQA Benchmark:} Tab~\ref{tab:vqa} shows the performance of BiMediX2 and other models on the Medical VQA benchmark using the MultiMedEval toolkit. Our BiMediX2 8B achieves the highest average score of 0.611, outperforming other models across datasets such as Rad-VQA, Slake-VQA, and Path-VQA. This demonstrates BiMediX2's proficiency in visual question answering, a critical task \mbox{in medical diagnostics.}

\noindent\textbf{Report Summarization:} Tab~\ref{tab:report_summary} presents the report summarization performance on the MIMIC-III dataset. BiMediX2 8B achieved the highest average score of 0.416, surpassing other models like LLaVA-Med (0.398) and Dragonfly-Med (0.110). This average score is derived as a unified metric by re-scaling BLUE-4* and RadCliQ* metrics. This indicates BiMediX2's effectiveness in generating concise and accurate summaries of medical reports, a vital task for efficient healthcare communication.

\noindent\textbf{Report Generation:} Tab~\ref{tab:report_generation} shows the report generation performance on MIMIC-CXR. BiMediX2 8B achieved an average score of 0.235, outperforming other models like LLaVA-Med (0.192) and BioMedGPT (0.145). The average score is derived as a unified metric by re-scaling BLUE-4* and RadCliQ* metrics. This highlights BiMediX2's capability in generating detailed and accurate medical reports from radiology images, a crucial task for diagnostic purposes. While MAIRA-2 † outperforms BiMediX2 on the Report Generation task, its performance is substantially lower on our comprehensive BiMed-MBench benchmark using GPT-4o (see Tab~\ref{tab:llavamed_eval_eng}, Tab~\ref{tab:llavamed_eval_ara}). This discrepancy is likely due to MAIRA-2's specialized fine-tuning for report generation as indicated in Tab~\ref{tab:comparison_llm}, whereas BiMediX2 maintains strong generalization across diverse multimodal biomedical tasks. 
For Med-PaLM M (562B), we report results directly from \cite{tu2023generalistbiomedicalai}, as the model is close-sourced and is not publicly available for direct evaluation. 

\section{Additional Experiments}
\noindent\textbf{Multi-stage Training Pipeline: }
To assess the impact of each component in our training framework, we evaluate our 8B model at three key stages. This staged evaluation allows us to disentangle the individual contributions of vision-language alignment and multimodal fine-tuning. \\ \textit{(i) Baseline}: VLM trained on non-medical Data. \textit{(ii) Stage-1}: Aligning Medical images to the LLM input token space by learning an alignment layer and freezing the Language model and the Vision encoder using our pre-training dataset of 467k instructions. \textit{(iii) Stage-2:} LoRA finetuning with our BiMed-V1.6M dataset comprising a text and image+text dataset across medical image modalities.

Here the Baseline model is obtained by pretraining the projector on LCS-558K dataset following the LLaVA-pp \cite{hanoona2024LLaVA++} repository. The baseline model lacks understanding of medical images which is introduced in our Stage-1 alignment training. Finally we LoRA finetune model to obtain our medically instruction tuned model. As observed in Tab~\ref{tab:multistage_ablation}, the text based medical evaluation scores do not change with Stage 1 training as we are only training the projector here and the language model performance is therefore consistent.

\noindent\textbf{Evaluation of English Model with Arabic Translation Pipeline:}
To further validate the necessity of bilingual instruction tuning, we employed a cascaded translation pipeline for evaluating our English instruction-tuned model (BiMediX2-8B ENG) on the BiMed-MBench Arabic benchmark using popular translation services such as from Google, Alibaba, and Bing.

\noindent While this method yielded slight improvements over the base English model as shown in Tab~\ref{tab:translation_exp}, it consistently fell short compared to our bilingual model. These results demonstrate that translation pipelines alone are inadequate for generating medically accurate Arabic responses, highlighting the importance of dedicated bilingual training and instruction tuning. Further experiments are detailed in Section \ref{sec:ablation}.

\begin{table}[!t]
\centering
\resizebox{0.9\linewidth}{!}{%
\begin{tabular}{@{}lc@{}}
\toprule
\textbf{Model}                                                                   & \multicolumn{1}{l}{\textbf{BiMed-MBench (Ara)}} \\ \midrule
BiMediX2-8B ENG                                                                  & 44.61                                           \\ \midrule
\begin{tabular}[c]{@{}l@{}}BiMediX2-8B ENG\\ + Translation (Google)\end{tabular} & 44.92                                           \\ \midrule
\begin{tabular}[c]{@{}l@{}}BiMediX2-8B ENG\\ + Translation (Bing)\end{tabular} & 45.05                                             \\ \midrule
\begin{tabular}[c]{@{}l@{}}BiMediX2-8B ENG\\ + Translation (Alibaba)\end{tabular} & 45.33                                           \\ \midrule
BiMediX2-8B BI                                                                   & 50.47                                           \\ \bottomrule
\end{tabular}
}
\caption{\textbf{Comparison of Translation frameworks}}
\label{tab:translation_exp}
\end{table}

\section{Qualitative Examples}

\textbf{Medical Image Understanding in a Conversational Context}:  Fig~\ref{fig:main_qualitative} illustrates the capabilities of the BiMediX2 framework in analyzing medical images and providing detailed, context-aware responses in both English and Arabic. The top section highlights BiMediX2 analyzing a sagittal CT scan of the lumbar spine, correctly identifying the scan type and diagnosing an L4 vertebral fracture. It explains potential causes like trauma or stress and discusses clinical implications. In the bottom section, BiMediX2 accurately identifies a female reproductive organ, explains the imaging technique, and detects abnormalities, demonstrating bilingual capabilities in English and Arabic. Additional qualitative examples are provided in Section \ref{sec:additional_qualitative}, while Section \ref{sec:failure_case} presents its limitations.

\section{Conclusion}

BiMediX2 represents a leap forward in bilingual, multimodal medical AI, addressing the global need for accessible and inclusive healthcare solutions in both English and Arabic. By integrating text and visual modalities within a unified architecture, it enables seamless multi-turn interactions for diverse medical tasks, including medical image analysis and complex medical conversations. Key contributions include the comprehensive bilingual dataset, BiMed-V, which provides diverse multimodal medical instructions tailored for both languages, and the introduction of BiMed-MBench, the first bilingual GPT-4o-based medical benchmark, which showcases the model's ability to excel in a wide range of expert-verified medical scenarios. BiMediX2 paves the way for inclusive, multilingual, and multimodal healthcare applications, significantly enhancing the accessibility and quality of \mbox{medical assistance worldwide.}

\begin{table}[!t]
\resizebox{\linewidth}{!}{%
\begin{tabular}{@{}lrrr@{}}
\toprule
\multicolumn{1}{c}{\textbf{Evaluation}} & \multicolumn{1}{l}{\textbf{Baseline}} & \multicolumn{1}{l}{\textbf{Stage-1}} & \multicolumn{1}{l}{\textbf{Stage-2}} \\ \midrule
\textbf{BiMed-MBench} [\ref{tab:llavamed_eval_eng}] & 34.9 & 47.6 & 62.2 \\
\textbf{Clinical LLM Eval} [\ref{tab:lmeval_sota}]  & 67.6 & 67.6 & 70.4 \\ \bottomrule
\end{tabular}%
}
\caption{\textbf{Multi-stage Training Ablation}}
\label{tab:multistage_ablation}
\end{table}

\section{Limitations}

Despite its overall improvement, BiMediX2, like other generative language models, may experience issues such as hallucinations, toxicity, and stereotypes. These issues stem from both the inherited limitations of the base models and the nature of the pretraining data. While we have conducted automatic and qualitative evaluations with medical professionals, we acknowledge that our model’s medical diagnoses and recommendations may not always be accurate. Extensive human evaluation is more reliable but expensive and time-consuming. The exploration of alternative solutions remains an important focus for ongoing research.

Currently, our models lack explicit mechanisms to curb undesirable behaviors. Future work will focus on enhancing alignment and safety strategies to reduce risks associated with clinical deployment. On a brighter note, we believe that releasing our model weights could contribute to investigating and mitigating these risks through broader community engagement. In addition, our current model does not explicitly mitigate biases related to gender, ethnicity, or socioeconomic status in medical contexts. We recognize the critical importance of fairness in AI-driven healthcare solutions, and acknowledge that such biases could perpetuate or exacerbate disparities in medical decision-making. As part of future work, we plan to conduct targeted bias assessments and explore debiasing strategies. We also believe that open-sourcing our model will facilitate further research into bias \mbox{detection and mitigation.}

Another current limitation is that our model is trained exclusively on Modern Standard Arabic (MSA). This restricts its applicability in real-world clinical settings where dialectal Arabic, such as Emirati, Saudi, Egyptian, or Moroccan is more commonly used. In future iterations of this work, we aim to expand our training corpus to include a diverse range of Arabic dialects, thereby enhancing the model’s utility and inclusivity across Arabic-speaking populations.

\section{Safety and Ethical Implications}

We recognize the significant societal impact of BiMediX2 and emphasize the importance of ethical considerations and transparency. This work is intended for research purposes only and is not ready for clinical or commercial use. Ensuring the model's accuracy and reliability is crucial, as incorrect medical advice could have serious health consequences. Robust validation and quality control measures are essential to minimize errors. 

Ethical considerations include protecting patient privacy and ensuring the confidentiality of medical data. The model must comply with relevant data protection regulations and ethical guidelines. Addressing potential biases in the model's outputs is also critical for ensuring fair and equitable \mbox{healthcare outcomes.}

Collaboration with patients, medical professionals, and ethicists is essential for ethical oversight and further research to ensure safety and accuracy in clinical settings. By acknowledging and addressing these considerations, we can continue to refine BiMediX2 for safe and effective use in healthcare. 

\section{Acknowledgement}
We would like to thank Dr.~Omair Mohammed, Dr.~Mohammed Zidan, Dr.~Vishal Thomas Oommen and Dr.~Mehrubin Kurpat for their contribution in verification of medical responses.

The computations were enabled by resources provided by \textit{LUMI hosted by CSC (Finland)} and \textit{LUMI consortium}, and by \textit{Berzelius} resource provided by the \textit{Knut and Alice Wallenberg \mbox{Foundation at the NSC}}.

This work is partially supported by the \textit{Meta Llama Impact Innovation Award}, the \textit{Meta Regional Research Grant}, the \textit{Google Research Award}, the \textit{NVIDIA Academic Grant}, and the \textit{MBZUAI-WIS Research Grant (P008)}. These recognitions highlight our commitment to advancing AI-driven healthcare solutions.

\vspace{1cm}

\bibliography{main}

\clearpage
\newpage

\appendix

\section{Appendix}
\label{sec:appendix}

\subsection{Additional Qualitative Examples}
\label{sec:additional_qualitative}

\begin{figure*}[!ht]
  \centering
    \includegraphics[width=\linewidth]{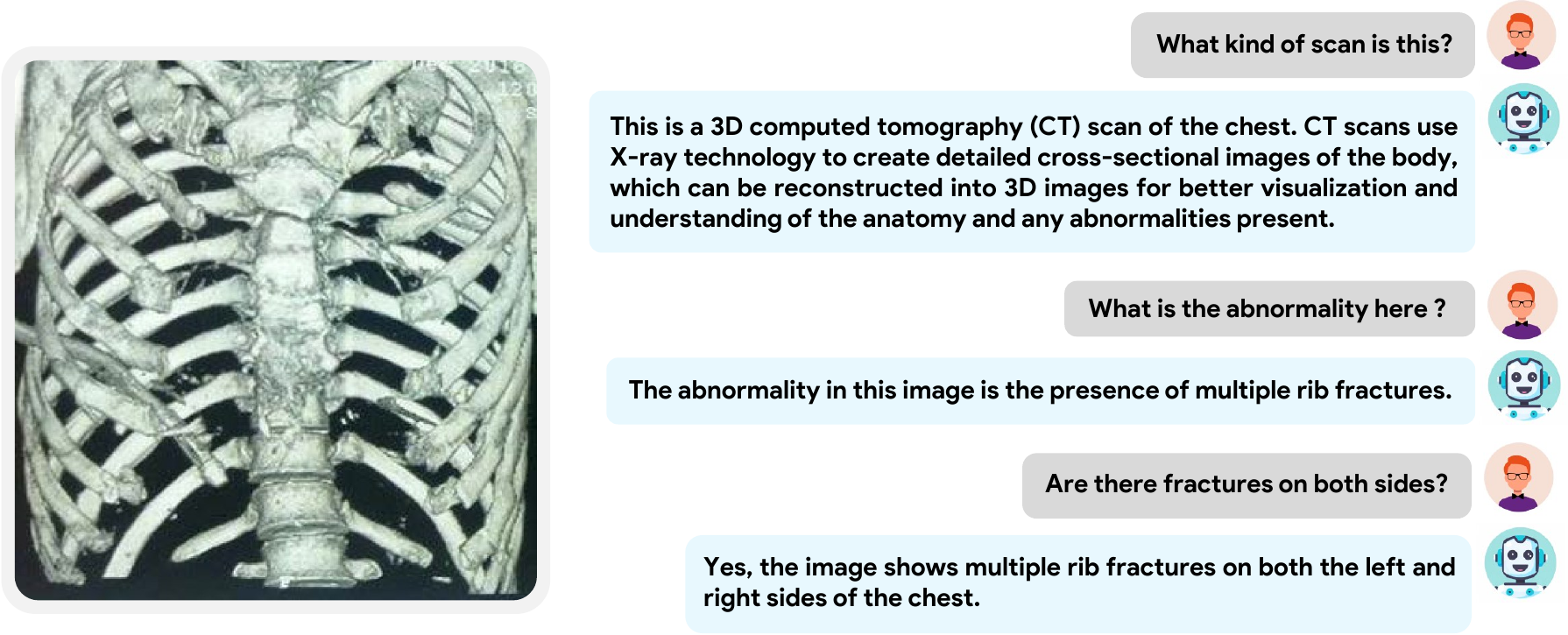}
    \includegraphics[width=0.02\linewidth]{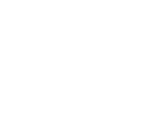}
    \includegraphics[width=\linewidth]{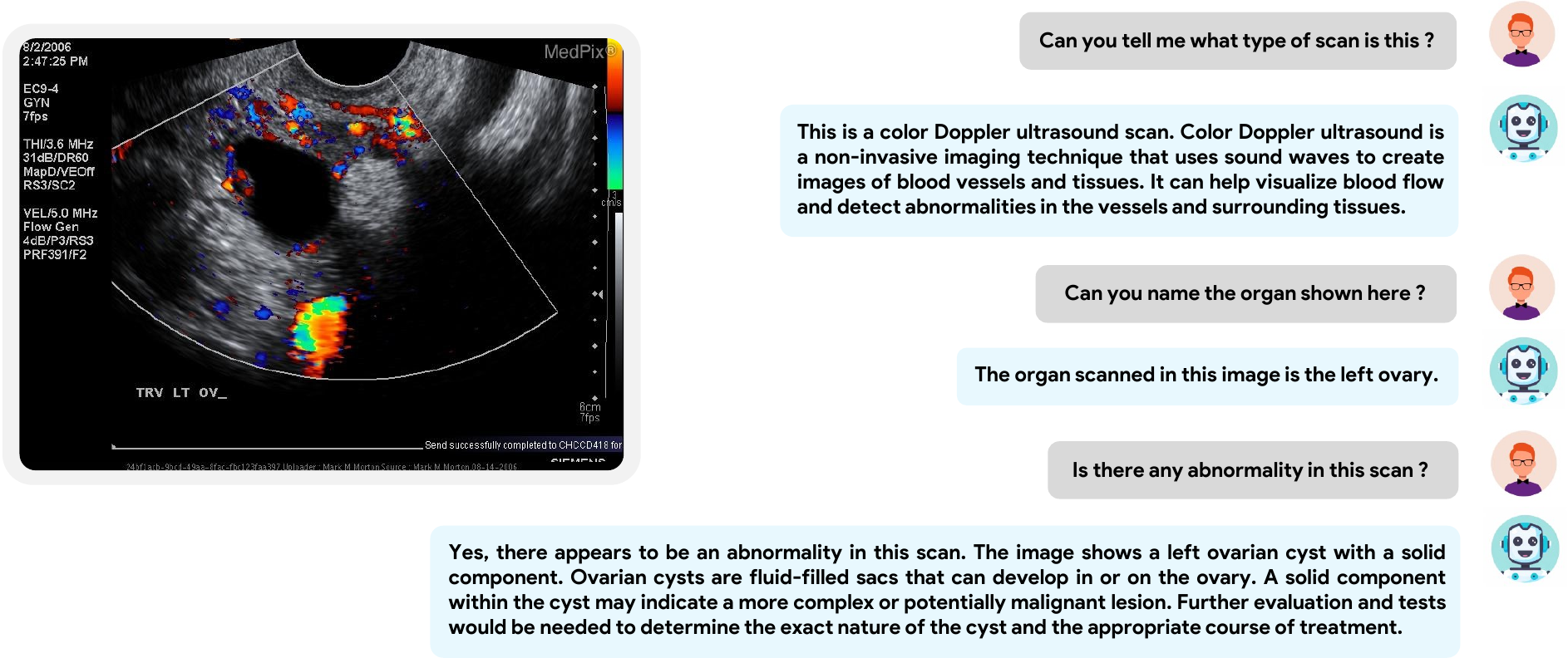}
    \caption{\textbf{Qualitative Examples of BiMediX2 for Medical Image Understanding in a Conversational Context.}}
    \label{fig:example1}
\end{figure*}

\textbf{Medical Image Understanding in a Conversational Context}:  Fig~\ref{fig:example1} illustrates the capabilities of the BiMediX2 framework in analyzing medical images and providing detailed, context-aware responses in both English and Arabic. The top section showcases a conversation involving a 3D CT scan of the chest. The model identifies the scan type and explains that it uses X-ray technology to create detailed cross-sectional images, which are then reconstructed into 3D images. When asked about abnormalities, the model accurately identifies multiple rib fractures.  It further clarifies that these fractures are present on both the left and right sides of the chest. And in the bottom section, BiMediX2 accurately identifies the organ, explains the imaging technique, and detects abnormalities, providing valuable insights that can aid in the diagnosis and treatment of conditions related to the female reproductive system. This particular example showcases BiMediX2's capability to converse in both English and Arabic simultaneously, depending on \mbox{the input query.}

\begin{figure*}[!ht]
  \centering
    \includegraphics[width=\linewidth]{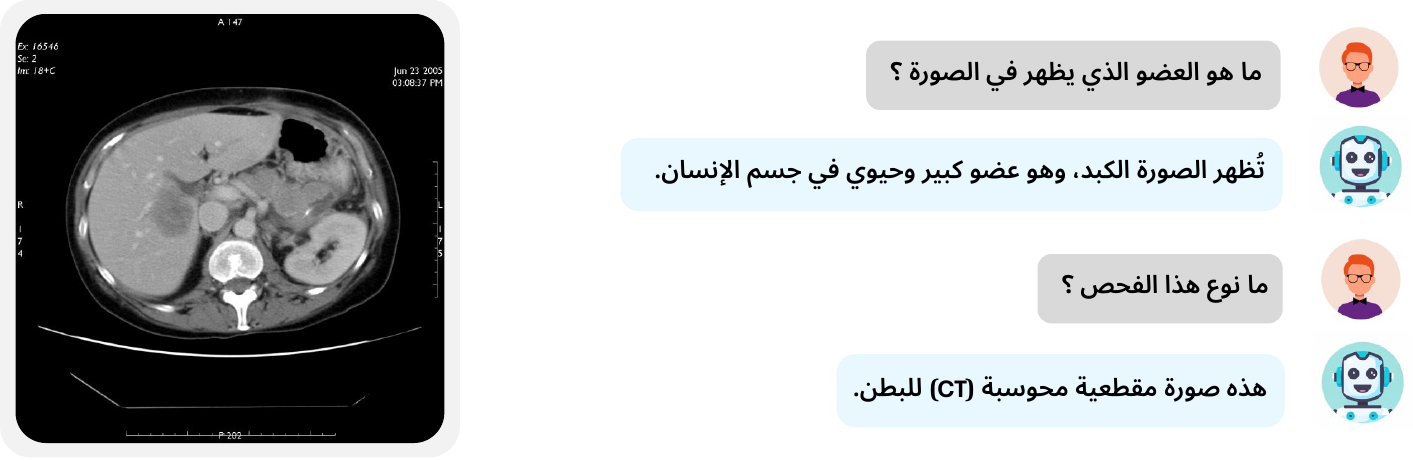}
    \includegraphics[width=0.02\linewidth]{Figures/white_space.png}
    \includegraphics[width=\linewidth]{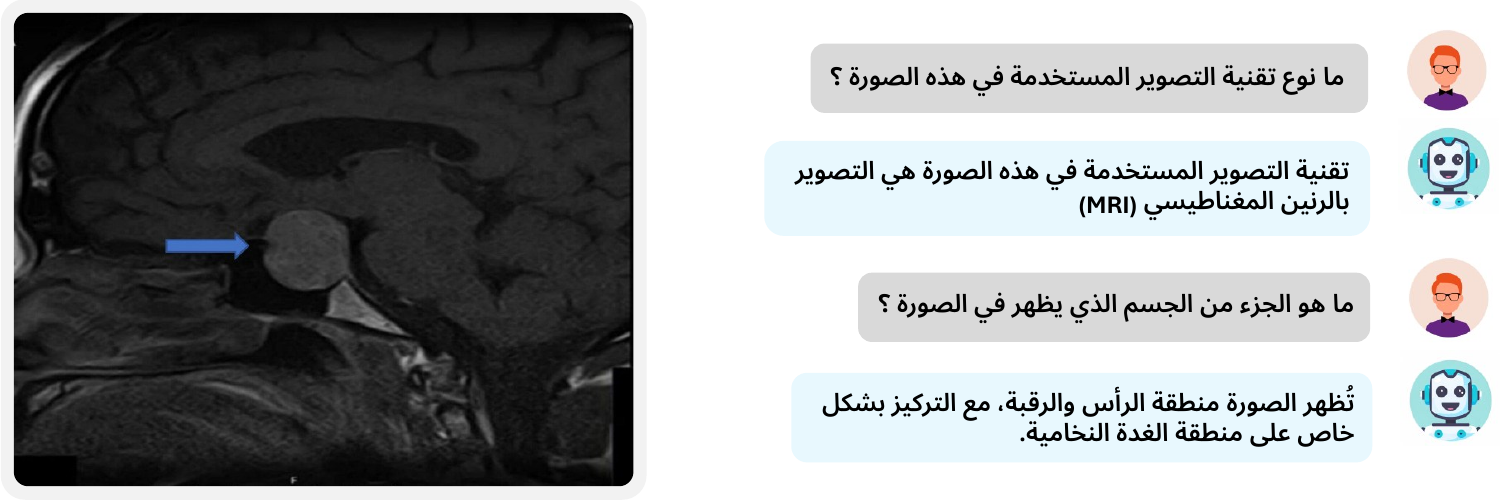}
    \includegraphics[width=0.02\linewidth]{Figures/white_space.png}
    \includegraphics[width=\linewidth]{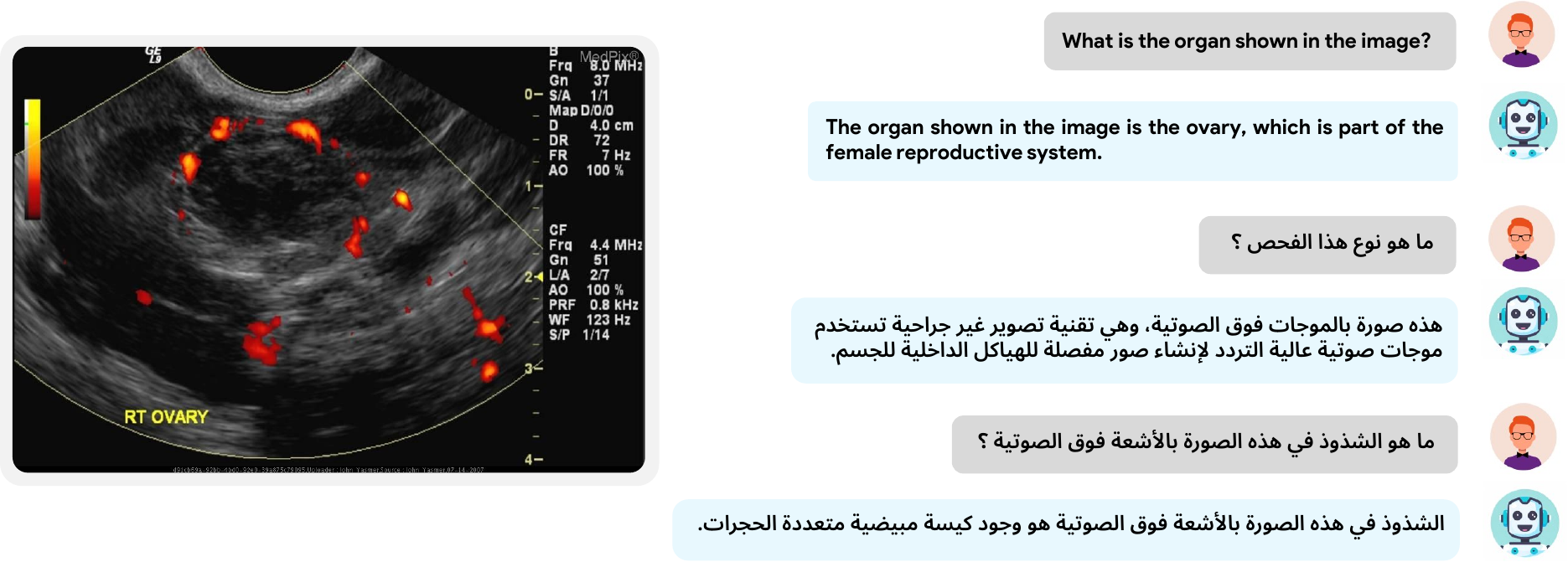}
    \caption{\textbf{Qualitative Examples of our BiMediX2 for Medical Image Understanding in a Conversational and Bilingual (Arabic) Setting:} The top section translates to; \textbf{User:} \textit{What is the organ shown in the image?} \textbf{BiMediX2:} \textit{The image shows the liver, which is a large and vital organ in the human body.} \textbf{User:} \textit{What type of scan is this?} \textbf{BiMediX2:} \textit{This is a computed tomography (CT) scan of the abdomen.} The middle section translates to; \textbf{User:} \textit{What type of imaging technique is used in this image?} \textbf{BiMediX2:} \textit{The imaging technique used in this image is magnetic resonance imaging (MRI).} \textbf{User:} \textit{What part of the body is shown in the image?} \textbf{BiMediX2:} \textit{The image shows the head and neck area, with a particular focus on the pituitary gland.} And the bottom section translates to; \textbf{User:} What is the organ shown in the image? \textbf{BiMediX2:} The organ shown in the image is the ovary, which is part of the female reproductive system. \textbf{User:} \textit{What type of scan is this?} \textbf{BiMediX2:} \textit{This is an ultrasound image, which is a non-invasive imaging technique that uses high-frequency sound waves to create detailed images of the body's internal structures.} \textbf{User:} \textit{What is the abnormality in this ultrasound image?} \textbf{BiMediX2:} \textit{The abnormality in this ultrasound image is the presence of a multilocular ovarian cyst.}}
    \label{fig:ara}
\end{figure*}

\noindent\textbf{Medical Image Understanding in a Conversational and Bilingual (Arabic) Setting}: Fig~\ref{fig:ara} illustrates BiMediX2's ability to understand medical imagery and converse in Arabic, showcasing its bilingual capabilities. In the top section, our model accurately identifies the organ and the type of scan, providing clear and precise information that can assist medical professionals in diagnosing and treating liver-related conditions. The middle section shows that our model identifies the body part and the type of scan. And in the bottom section, BiMediX2 accurately identifies the organ, explains the imaging technique, and detects abnormalities, providing valuable insights that can aid in the diagnosis and treatment of conditions related to the female reproductive system. This particular example showcases BiMediX2's capability to converse in both English and Arabic simultaneously.

\begin{figure*}[!htp]
  \centering
    \includegraphics[width=0.97\linewidth]{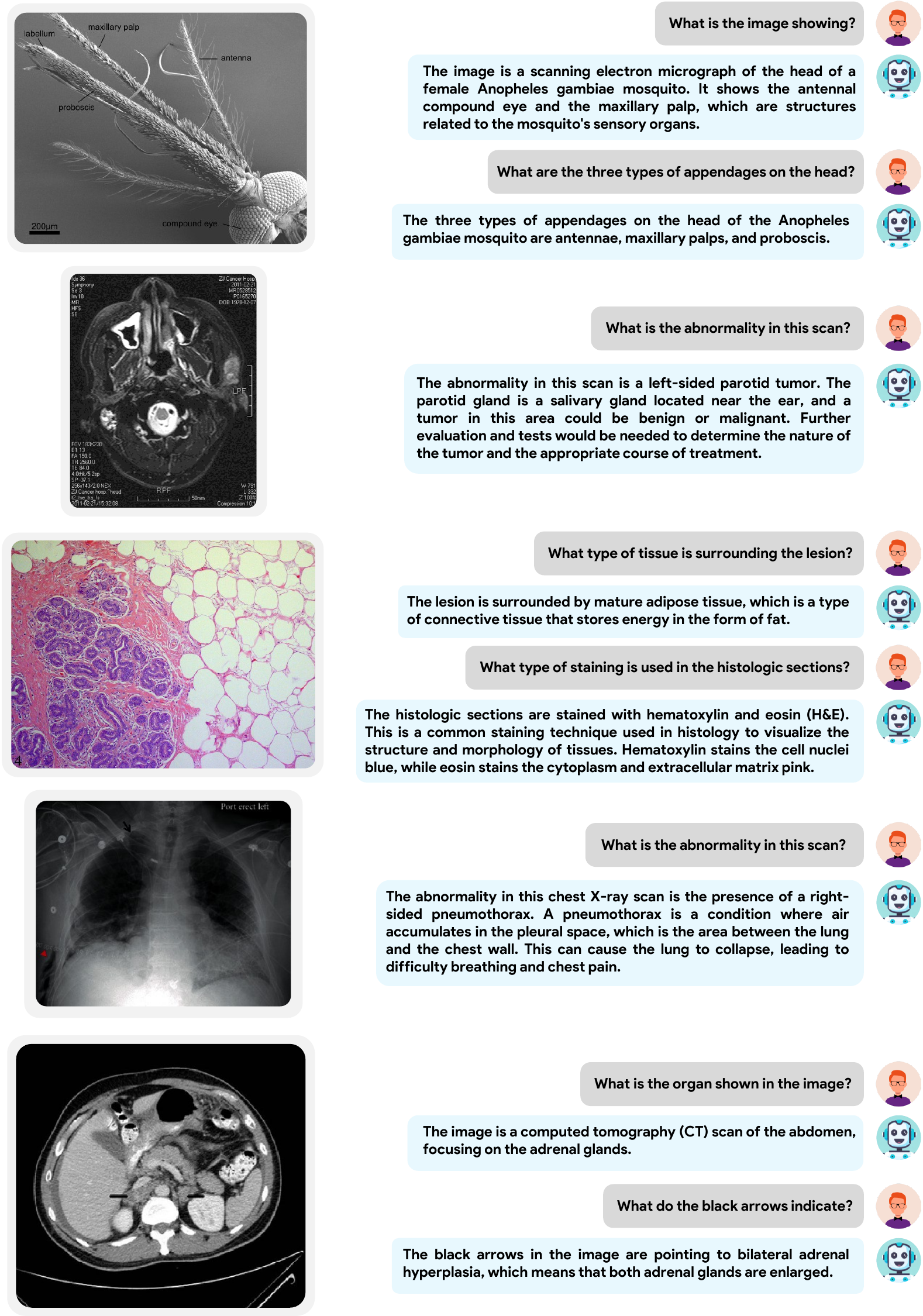}
    \caption{\small{\textbf{Qualitative Examples for Medical Image Understanding of our BiMediX2 in Multidomain}: Capability of BiMediX2 to understand and interpret various medical imaging modalities and provide accurate responses. The examples include describing the anatomy of a mosquito's head in a scanning electron micrograph, detecting a left-sided parotid tumor in an MRI scan, recognizing mature adipose tissue in a histology slide, identifying a right-sided pneumothorax in a chest X-ray, and identifying bilateral adrenal hyperplasia in a CT scan of the abdomen. These examples highlight BiMediX2's versatility and effectiveness in medical image analysis and diagnosis.}}
    \label{fig:multidomain}
\end{figure*}

\noindent\textbf{Medical Image Understanding of our BiMediX2 in Multidomain}: Fig~\ref{fig:multidomain} showcases BiMediX2's versatility and accuracy across various medical imaging modalities. In the top section, it identifies key structures in a scanning electron micrograph of a mosquito head. In subsequent sections, it correctly interprets an MRI of a parotid tumor, a histology slide of adipose tissue (including stain type), a chest X-ray with pneumothorax, and a CT scan of the abdomen, pinpointing adrenal abnormalities. These examples demonstrate BiMediX2's capability to analyze and diagnose diverse imaging types, making it a valuable tool across \mbox{medical specialties.}

\begin{figure*}[!ht]
  \centering
    \includegraphics[width=\linewidth]{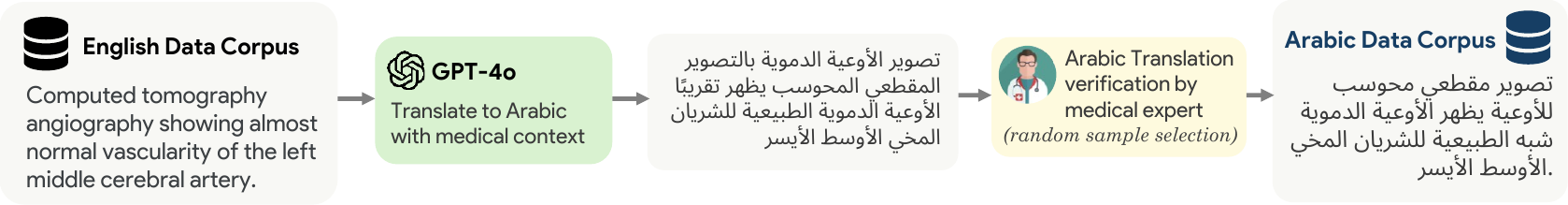}
    \caption{\textbf{Data Translation Framework}}
    \label{fig:data_translation}
\end{figure*}

\subsection{Model Failure Cases}
\label{sec:failure_case}
\begin{figure*}[!ht]
  \centering
    \includegraphics[width=\linewidth]{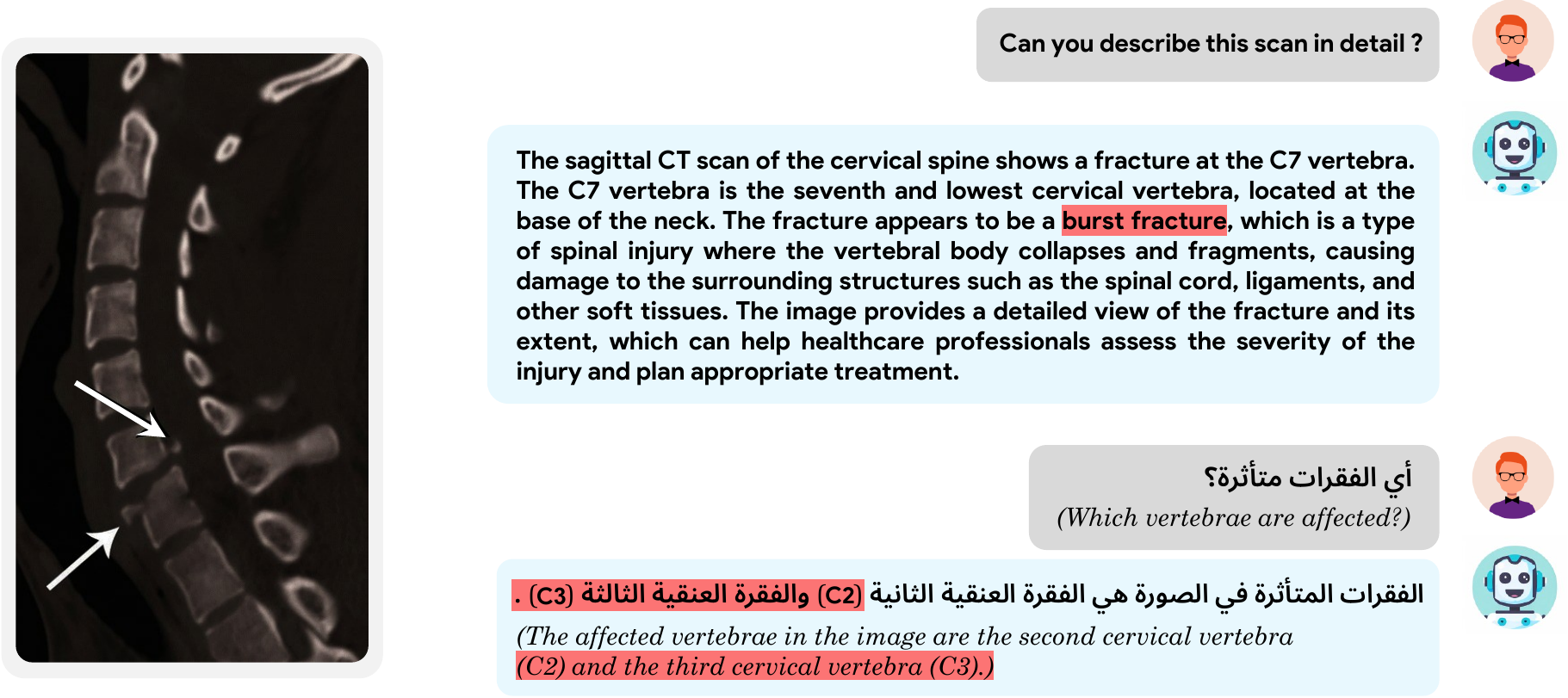}
    \caption{\textbf{Bilingual failure case:} Model labels avulsion fracture as burst (English), and misidentifies C2–C3 instead of C6–C7 (Arabic).}
    \label{fig:error_analysis}
\end{figure*}

We present a common failure case observed in both English and Arabic settings. The example is \mbox{shown in Fig~\ref{fig:error_analysis}.}

In the English-language evaluation, a common source of error was the model's difficulty in distinguishing between visually similar but clinically distinct fracture types. As shown in Fig~\ref{fig:error_analysis}, while the model response is fluent and medically coherent, it contains a key factual error: the misclassification of an avulsion fracture as a burst fracture. Notably, this response was initially accepted by several medical professionals due to the similarity in language and presentation. However, a trained radiologist later identified the subtle radiographic distinctions, confirming the ground truth diagnosis as an avulsion fracture. This highlights the model's challenge in handling nuanced diagnostic distinctions that require expert-level domain knowledge.

In the Arabic benchmark, one of the most frequent error modes involved incorrect localization of the affected area as illustrated in Fig~\ref{fig:error_analysis}. The response demonstrates a localization mismatch where the model incorrectly identifies C2–C3 as the affected vertebrae instead of the correct C6–C7 levels. Such errors, especially in high-stakes medical applications, can have serious clinical implications, reinforcing the importance of precise anatomical grounding in medical AI systems.

\subsection{Clinical LLM Benchmarks}
\label{sec:llm_bench}

\textbf{PubMedQA} \cite{jin2019pubmedqa} is a question-answering dataset derived from biomedical research papers on PubMed. The task involves answering 'yes', 'no', or 'maybe' based on question sourced from the title of a research paper and a context from the abstract. Our analysis focuses on the PQA-L subset, which includes 500 manually annotated QA pairs requiring in-depth \mbox{biomedical reasoning.}

\noindent\textbf{MedMCQA} \cite{pal2022medmcqa} consists of 4,183 multiple-choice questions from Indian AIIMS and NEET PG medical entrance exams, assessing professional medical knowledge and \mbox{language comprehension.}

\noindent\textbf{MedQA} \cite{jin2021disease} features multiple-choice questions from medical board exams in the US, Mainland China, and Taiwan. Our study focuses on the USMLE portion (1,273 test samples), requiring multi-step reasoning \mbox{and evidence retrieval.}

\noindent\textbf{USMLE} \cite{han2023medalpaca} is a self-assessment test from the United States Medical Licensing Examination (Step 1, Step 2, and Step 3). We use the MedAlpaca version, which excludes image-based questions and contains 325 test samples.

\noindent\textbf{Medical MMLU} \cite{hendrycks2020measuring} is a collection of six datasets covering 1,089 test questions across Clinical Knowledge, College Biology, College Medicine, Medical Genetics, Professional Medicine, and Anatomy.

\noindent\textbf{UPHILL OpenQA} \cite{kaur2023evaluating} evaluates LLMs' accuracy in handling health-related queries with varying presuppositions. Our analysis focuses on zero-shot models' ability to refute false health claims, a critical factor in combating misinformation. In this context, the accuracy refers to the model’s effectiveness in accurately refuting false health-related claims at \mbox{different presupposition levels.}

\begin{figure*}[!ht]
  \centering
    \includegraphics[width=\linewidth]{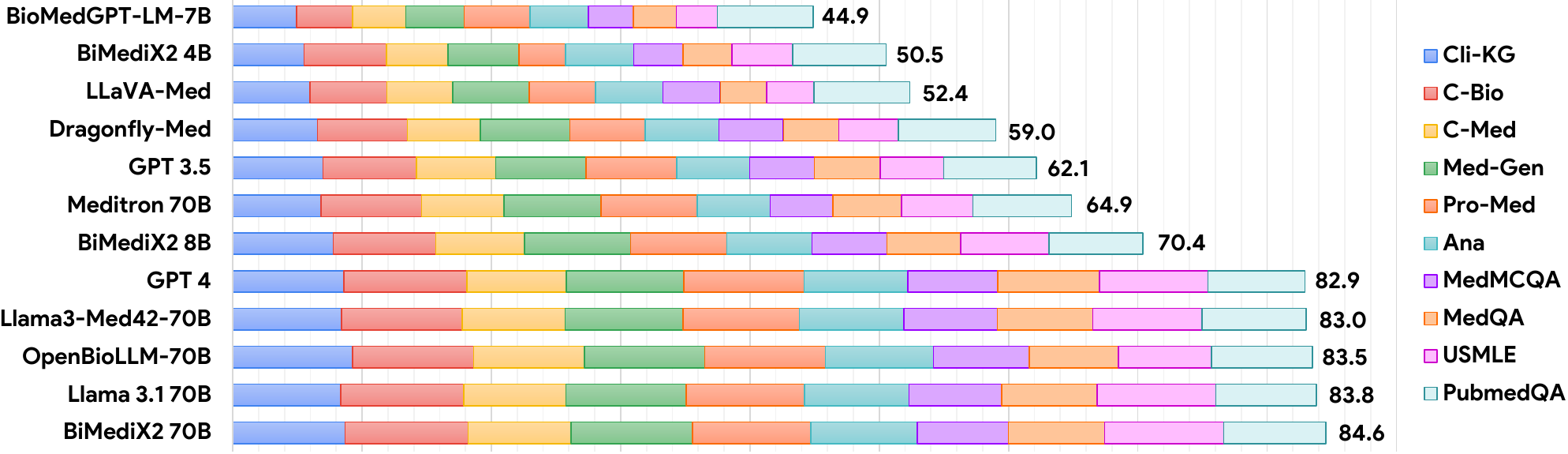}
    \caption{\textbf{State of the art comparison of models in Clinical LLM Benchmarks}}
    \label{fig:lmeval_sota}
\end{figure*}

Fig ~\ref{fig:lmeval_sota} compares state-of-the-art medical LLMs and LMMs on clinical benchmarks. BiMediX2 70B achieved the highest average score (84.6\%), surpassing GPT-4 (82.9\%) and Llama-3-Med42-70B (83.0\%), demonstrating its superior performance across diverse medical tasks.  The stacked bars illustrate performance across individual datasets, highlighting BiMediX2's strong understanding of medical contexts.

\subsection{Data Translation Framework}
\label{sec:data_translation}

To construct a high-quality bilingual dataset, we develop a robust data translation framework that leverages GPT-4o for translating English medical content into Arabic, followed by expert human verification to ensure contextual and terminological accuracy. This approach supports effective training and benchmarking in a bilingual context.

We first translate our English multimodal instruction set to Arabic using GPT-4o. A random subset of this translated data is passed to Bilingual medical doctors to verify the quality of the translation. Doctors are asked to report the number of samples requiring minor (e.g rewording, formatting, or stylistic edits) / major (e.g incorrect clinical terms or misinterpretations) revision in the Arabic translation. They are required to look for the overall consistency of the translated text and the accuracy of medical terminologies. 

For our BiMed-MBench bilingual benchmark doctors were asked to manually correct these inaccuracies in the translated text. On average 22\% of the samples required minor corrections and re-formatting, while only 5\% of the samples required major corrections in medical terms.

The verification process involved 10 medical experts from three different countries, ensuring representation from both native Arabic and English speakers. These experts specialized in various medical domains, including radiology (MD Radiology), endocrinology, neurosurgery, general practice, histopathology (MS Pathology), and general medicine. To ensure rigorous validation, a multi-reviewer setup was adopted for a randomly selected subset of the benchmark. Each sample was independently reviewed by multiple doctors, which surfaced occasional inconsistencies, particularly in ambiguous or borderline cases.

To resolve such discrepancies, we introduced a structured adjudication protocol. Conflicting evaluations were circulated among the experts, followed by focused discussion sessions in which reviewers examined each other’s rationale. This collaborative refinement process significantly reduced inconsistencies and improved the reliability \mbox{of the benchmark.}

\subsection{Additional Experiments}
\label{sec:ablation}

\begin{table}[!ht]
\centering
\resizebox{0.8\linewidth}{!}{%
\begin{tabular}{@{}lr@{}}
\toprule
\multicolumn{1}{c}{\textbf{Model}} & \multicolumn{1}{l}{\textbf{BiMed-MBench (Ara)}} \\ \midrule
\textbf{BiMediX2-8B ENG} & 44.6 \\
\textbf{BiMediX2-8B ARA} & 46.3 \\
\textbf{BiMediX2-8B BI} & 50.5 \\ \bottomrule
\end{tabular}%
}
\caption{\textbf{Bilingual vs Monolingual model training}}
\label{tab:bi_vs_mono}
\end{table}

\begin{table}[!t]
\centering
\resizebox{\linewidth}{!}{%
\begin{tabular}{@{}lll@{}}
\toprule
\multicolumn{1}{c}{\textbf{Hyperparameter}} & \multicolumn{1}{c}{\textbf{Stage 1}} & \multicolumn{1}{c}{\textbf{Stage 2}} \\ \midrule
Number of Epochs                            & 1                                    & 1                                    \\
Train Batch Size (per device)               & 1                                    & 4                                    \\
Gradient Accumulation Steps                 & 1                                    & 4                                    \\
Learning Rate                               & 1.0×10\textsuperscript{-3}           & 2.0×10\textsuperscript{-4}           \\
Optimizer                                   & Adam                                 & Adam                                 \\
Weight Decay                                & 0                                    & 0                                    \\
Warmup Ratio                                & 0.03                                 & 0.03                                 \\
Learning Rate Scheduler                     & Cosine                               & Cosine                               \\
Precision                                   & bfloat16                             & bfloat16                             \\
PEFT                                        & None                                 & LoRA                                 \\
LoRA Rank                                   & -                                    & 8                                    \\
LoRA Alpha                                  & -                                    & 16                                   \\
Multi-Modal Projector LR                    & -                                    & 2.0×10\textsuperscript{-5}           \\ \bottomrule             
\end{tabular}
}
\caption{\textbf{Model training Hyperparameters}}
\label{tab:train_spec}
\end{table}

\subsubsection{Human Expert Evaluation}

To assess the clinical quality of model outputs, we conducted a human expert evaluation using a blind review setup. For a randomly selected subset of questions from the BiMed-MBench benchmark, responses were generated by three models: BiMediX2, Dragonfly-Med, and LLaVA-Med. The outputs were anonymized and labeled as Model A, Model B, and Model C, with no identifiers provided to the reviewers.

Medical experts were asked to evaluate the responses against the provided ground truth descriptions for each question. The evaluation focused on determining which model produced the most accurate, clinically relevant, and clear explanation of the medical image.

The results demonstrate a strong preference for BiMediX2, which was selected as the best response in 76.9\% of the cases. In comparison, Dragonfly-Med was preferred in 15.4\% of the cases, and LLaVA-Med in 7.7\%. \\

\noindent\textbf{Evaluation Protocol for Medical Experts:} \\ 
\begin{small}
\noindent Your task is to evaluate the responses provided by three AI models based on a given medical image description (Ground Truth). Follow these steps to make your selection:

\noindent 1) Read the Ground Truth: Carefully review the provided description of the medical image. This serves as the reference for an accurate and detailed response.

\noindent 2) Assess the Model Responses: Examine the three model-generated responses (Model A, Model B, and Model C). Compare their content with the Ground Truth, focusing on the accuracy, completeness, and relevance of the description.

\noindent 3) Select the Best Response: Choose the model response that best aligns with the Ground Truth in terms of:

\noindent - Clinical Accuracy: Does the response correctly describe the key findings in the image?

\noindent - Relevance: Does the response stay focused on the specific details highlighted in the Ground Truth?

\noindent - Clarity: Is the explanation well-structured and easy to \mbox{understand}

\noindent 4) Submit Your Choice: After evaluating the responses, select the one that provides the most accurate and comprehensive explanation.

\end{small}

\begin{table}[!t]
\centering
\resizebox{0.85\linewidth}{!}{%
\begin{tabular}{@{}ll@{}}
\toprule
\textbf{Dataset}                       & \textbf{No. of samples} \\ \midrule
PubMedQA                      & 210169         \\
MedMCQA                       & 182712         \\
MedQA                         & 20691          \\
Single Turn QA                & 119879         \\
Multi-Turn Conversation       & 133134         \\
PMC-MCQ                       & 80000          \\
LLaVA-Med-Subset-to-Conv      & 11616          \\
Rad-VQA                       & 1796           \\
Slake-VQA                     & 9835           \\
PMC-VQA                       & 80000          \\
Path-VQA                      & 19654          \\
LLaVA-Med-to-QA               & 163463         \\
MedQA (Ara)                   & 11210          \\
PubMedQA (Ara)                & 115773         \\
MedMCQA (Ara)                 & 97523          \\
Single Turn QA (Ara)          & 156254         \\
Multi-Turn Conversation (Ara) & 64235          \\
LLaVA-Med-to-QA (Ara)         & 163463         \\
PMC-MCQ (Ara)                 & 50000          \\ \midrule
\rowcolor{yellow!20}\textbf{Total}    & \textbf{1691407}  \\ \bottomrule
\end{tabular}
}
\caption{\textbf{BiMed-V-1.6M Dataset Composition}}
\label{tab:dataset_comp}
\end{table}

\subsubsection{Comparison of Bilingual vs Monolingual Training}

The results in Tab~\ref{tab:bi_vs_mono} highlight the impact of language setting on model performance for Arabic medical evaluation tasks. The bilingual model (BiMediX2-8B BI), trained on both English and Arabic instruction data, achieves the highest score of 50.5, outperforming both the Arabic-only model (46.3) and the English-only model (44.6) on the Arabic subset of BiMed-MBench.

This demonstrates that bilingual instruction tuning provides complementary knowledge transfer, enabling the model to better generalize in Arabic. The performance gain over the Arabic-only model further emphasizes the value of incorporating English medical knowledge during training, even when the target evaluation is in Arabic.

\subsection{Model Configurations}
\label{sec:model_config}

All traning experiments were conducted with 8× AMD Instinct MI200 GPUs (each with 64 GB of VRAM). The training process is structured in two stages: Stage 1 training requires approximately 20 hours, and Stage 2 training requires approximately 32 hours on this setup. The complete set of training hyperparameters are provided in Tab~\ref{tab:train_spec}.

We also compare the trade-offs between our model sizes along with their corresponding performance and medical accuracy in Tab~\ref{tab:model_variants}. We compare the Token Throughput measured in (Tokens/second), Mean Time to First Token (TTFT) (ms), GPU VRAM consumed and Medical Accuracy based on our evaluation benchmarks. These benchmarks are run using the vLLM benchmarking suite \cite{kwon2023efficient} on 4xNVIDIA RTX A6000 (48GB) GPUs.

\subsection{BiMedV-1.6M Dataset}
\label{sec:dataset_comp}

\begin{figure*}[!t]
  \centering
    \includegraphics[scale=0.5]{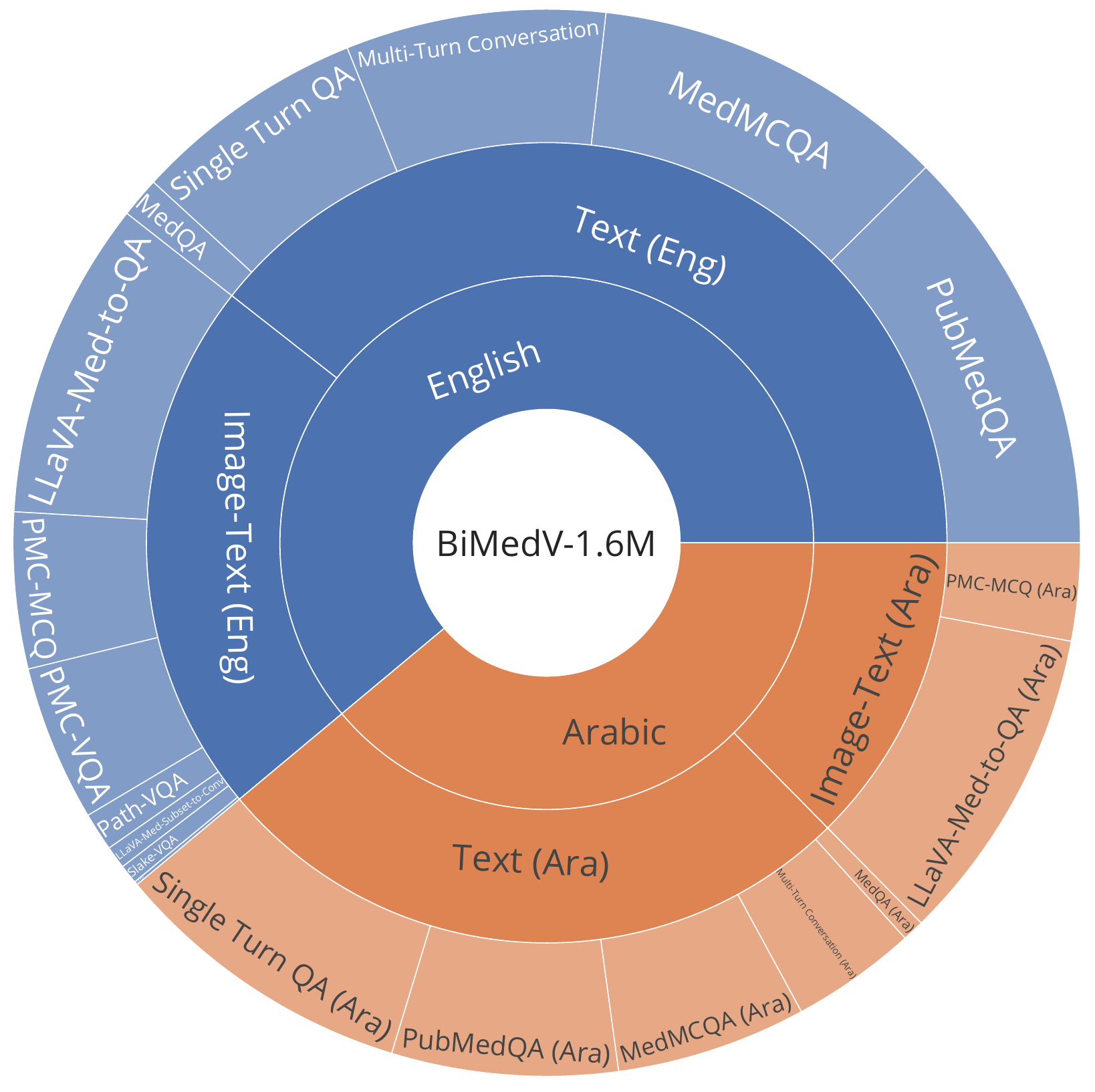}
    \caption{\textbf{BiMed-V-1.6M Dataset Distribution.} The sunburst chart illustrates the hierarchical composition of the dataset, categorizing samples by language (English and Arabic), modality (text and image-text), and task type (QA and VQA). The inner rings represent language and modality divisions, while the outer rings detail specific datasets included in BiMedV-1.6M.}
    \label{fig:dataset_chart}
\end{figure*}

The \textbf{BiMedV-1.6M dataset} comprises \textbf{1,691,407 samples} across English and Arabic, covering text and image-text QA tasks, supporting multilingual and multimodal medical research. The dataset composition is presented in Tab~\ref{tab:dataset_comp}, Fig~\ref{fig:dataset_chart}.

\begin{table*}[!ht]
\centering
\resizebox{0.95\textwidth}{!}{%
\begin{tabular}{@{}lrrrr@{}}
\toprule
\multicolumn{1}{c}{\textbf{Model}} & \multicolumn{1}{l}{\textbf{Token Throughput (tok/s)}} & \multicolumn{1}{l}{\textbf{Mean TTFT (ms)}} & \multicolumn{1}{l}{\textbf{GPU VRAM (GB)}} & \multicolumn{1}{l}{\textbf{Medical Accuracy}} \\ \midrule
\textbf{BiMediX2-4B} & 263.05 & 180.5 & 42 & 50.5 \\
\textbf{BiMediX2-8B} & 187.14 & 241.9 & 44 & 70.4 \\
\textbf{BiMediX2-70B} & 67.61 & 863.6 & 170 & 84.6 \\ \bottomrule
\end{tabular}%
}
\caption{\textbf{Performance comparison of BiMediX2 variants}}
\label{tab:model_variants}
\end{table*}

\end{document}